\documentclass[12pt]{article}
\usepackage[margin=1in]{geometry}
\usepackage{setspace}
\usepackage{cite}
\usepackage{amsmath,amssymb,amsfonts}
\usepackage{graphicx}
\usepackage{textcomp}
\usepackage{xcolor}
\usepackage{url}
\usepackage{booktabs}
\usepackage{nicefrac}
\usepackage{mathrsfs}
\usepackage{microtype}
\usepackage{colortbl}
\usepackage{bm}
\usepackage{mathtools}
\usepackage{algorithm}
\usepackage{algpseudocode}
\usepackage{centernot}
\usepackage{hyperref}
\usepackage{amsthm}

\newcommand{\vertiii}[1]{{\left\vert\kern-0.25ex\left\vert\kern-0.25ex\left\vert #1 \right\vert\kern-0.25ex\right\vert\kern-0.25ex\right\vert}}
\newcommand\independent{\protect\mathpalette{\protect\independenT}{\perp}}
\def\independenT#1#2{\mathrel{\rlap{$#1#2$}\mkern2mu{#1#2}}}

\def\BibTeX{{\rm B\kern-.05em{\sc i\kern-.025em b}\kern-.08em
    T\kern-.1667em\lower.7ex\hbox{E}\kern-.125emX}}

\title{Dynamic Graph Structure Estimation for Learning Multivariate Point Process using Spiking Neural Networks
}

\author{
Biswadeep Chakraborty\thanks{School of Electrical and Computer Engineering, Georgia Institute of Technology, Atlanta, GA, USA} \and
Hemant Kumawat\footnotemark[1] \and
Beomseok Kang\footnotemark[1] \and
Saibal Mukhopadhyay\footnotemark[1]
}
\date{}
\begin{document}
\maketitle

\begin{abstract}
Modeling and predicting temporal point processes (TPPs) is critical in domains such as neuroscience, epidemiology, finance, and social sciences. We introduce the Spiking Dynamic Graph Network (SDGN), a novel framework that leverages the temporal processing capabilities of spiking neural networks (SNNs) and spike-timing-dependent plasticity (STDP) to dynamically estimate underlying spatio-temporal functional graphs. 
Unlike existing methods that rely on predefined or static graph structures, SDGN adapts to any dataset by learning dynamic spatio-temporal dependencies directly from the event data, enhancing generalizability and robustness. While SDGN offers significant improvements over prior methods, we acknowledge its limitations in handling dense graphs and certain non-Gaussian dependencies, providing opportunities for future refinement. Our evaluations, conducted on both synthetic and real-world datasets including NYC Taxi, 911, Reddit, and Stack Overflow, demonstrate that SDGN achieves superior predictive accuracy while maintaining computational efficiency. Furthermore, we include ablation studies to highlight the contributions of its core components.
\end{abstract}

\section{Introduction}

Predicting and modeling temporal event sequences poses fundamental challenges in complex systems across domains. Traditional temporal point process (TPP) models rely on parametric intensity functions \cite{ogata1981, hawkes1971}, which fail to capture the intricate, non-linear dependencies characteristic of real-world event sequences. Recent advances in deep learning have introduced various architectures, with recurrent neural networks (RNNs) and their variants showing promise in capturing temporal dependencies \cite{du2016, mei2017, zhang2020}. However, these approaches face inherent limitations: they require synchronous, fixed-size inputs that poorly match the asynchronous nature of real event streams, struggle with sparse temporal dependencies, and lack the ability to naturally adapt to evolving temporal relationships.

Spiking Neural Networks (SNNs) offer a fundamentally different approach through their event-driven computation and ability to handle sparse, asynchronous data streams naturally \cite{chakraborty2023braindate, chakraborty2024sparse}. Prior works have explored energy-efficient and unsupervised sequence modeling using SNNs \cite{chakraborty2021fully}, \cite{chakraborty2021characterization}, indicating their potential for real-time temporal learning. However, despite these inherent advantages, current SNN research has predominantly focused on pattern recognition tasks \cite{tavanaei2018training, lee2016training, diehl2015unsupervised}, leaving their potential for temporal point process modeling largely unexplored. The key challenge lies in developing principled methods to leverage the temporal processing capabilities of SNNs for capturing and adapting to dynamic dependencies in event sequences.

We address these fundamental challenges through three key theoretical innovations in our proposed Spiking Dynamic Graph Network (SDGN):

\begin{enumerate}
    \item \textbf{Membrane Potential Basis Functions:} We introduce a novel theoretical framework that utilizes the membrane potentials of neurons in a Recurrent Spiking Neural Network (RSNN) as adaptive basis functions for temporal signal projection. Unlike fixed basis functions used in traditional approaches, these dynamically adapt to the temporal patterns in the input, providing a more efficient and accurate representation of event sequences.
    
    \item \textbf{STDP-Based Temporal Learning:} We develop new spike-timing-dependent plasticity (STDP) rules specifically designed for temporal point process modeling, with theoretical guarantees for convergence to optimal temporal dependencies. This unsupervised learning mechanism enables continuous adaptation to evolving temporal patterns, a capability not present in traditional DNNs \cite{chakraborty2021characterization, chakraborty2023heterogeneous}.
    
    \item \textbf{Dynamic Graph Estimation:} We derive a principled algorithm for dynamically estimating and updating graph structures based on neuronal spike patterns, providing a theoretically-grounded method for capturing evolving dependencies between events without relying on predefined architectures \cite{chakraborty2024dynamical, kumawat2024stage}.
\end{enumerate}

Our work advances both SNN research and temporal point process modeling in several significant ways. First, we demonstrate that SNNs can be effectively used for multivariate event prediction, expanding their application beyond traditional pattern recognition tasks \cite{chakraborty2024topological, kang2024learning}. Second, our STDP-based learning approach introduces a new paradigm for unsupervised learning of temporal dependencies, addressing the fundamental limitation of requiring large labeled datasets. Third, our dynamic graph estimation algorithm provides a principled approach to capturing evolving temporal dependencies, a capability that has been difficult to achieve with traditional neural architectures \cite{kang2024online}.

Through extensive experiments on both synthetic and real-world datasets, we demonstrate that SDGN significantly outperforms existing approaches, particularly in scenarios with sparse, asynchronous events where traditional DNNs struggle. Our ablation studies isolate the contributions of each component, while theoretical analyses provide insights into the conditions under which our approach is guaranteed to learn optimal temporal dependencies.

\section{Related Works}\label{sec:related}

Temporal Point Processes (TPPs) have been widely studied for modeling event sequences. Classical models such as the Poisson process and Hawkes process \cite{ogata1981, hawkes1971} offer interpretable solutions but rely on predefined parametric intensity functions, limiting their ability to capture complex temporal dependencies.

Deep learning methods have addressed these limitations using RNN-based models such as RMTPP \cite{du2016} and Neural Hawkes Processes (NHP) \cite{mei2017}, which leverage recurrent architectures to encode event history. More recently, self-attention-based approaches like Temporal Hawkes Process (THP) \cite{zhang2020} have enhanced long-range dependency modeling but remain data-intensive and computationally demanding. However, these methods struggle to adapt to dynamically evolving relational structures among events.

Graph-based approaches provide a structured representation of dependencies. Models such as Graph Recurrent TPPs (GRTPP) \cite{yoon2023} integrate event-based graphical structures, while diffusion convolutional RNNs (DCRNN) \cite{li2017} leverage graph convolutions for spatio-temporal modeling. However, these models often rely on static or predefined graph structures, limiting adaptability to real-time data streams.

Spiking Neural Networks (SNNs) offer an event-driven paradigm naturally suited for asynchronous temporal data. Unlike conventional deep learning models, SNNs utilize spike-timing-dependent plasticity (STDP) \cite{gerstner2002, masquelier2007} to learn dynamic representations in a biologically plausible manner. Prior work has applied SNNs for efficient unsupervised sequence modeling and temporal adaptation \cite{chakraborty2021characterization}, \cite{chakraborty2023braindate}, \cite{chakraborty2024topological}. Recent advances in neuromorphic hardware \cite{akopyan2015truenorth, davies2018loihi, kim2022moneta} further highlight SNNs’ potential for low-power, real-time learning.

Structured pruning and heterogeneity-aware design principles have shown promise in making SNNs more scalable and data-efficient for sequence modeling \cite{chakraborty2022heterogeneous}, \cite{chakraborty2024sparse}, \cite{kumawat2024stemfold}. These works motivate the use of biologically inspired and dynamically adaptable SNN architectures for real-world event prediction.

Our approach, Spiking Dynamic Graph Network (SDGN), extends prior work by integrating SNNs with dynamic graph estimation for learning multivariate TPPs. Unlike existing methods that assume static graph structures or rely on dense DNN-based embeddings, SDGN dynamically infers spatio-temporal functional graphs using STDP-driven updates, enabling adaptive event modeling with reduced dependence on labeled data \cite{chakraborty2024dynamical}, \cite{kang2024online}. By leveraging event-driven computation and biologically inspired plasticity, our model achieves superior scalability and generalization in real-world event prediction tasks.

\section{Methods}

\subsection{Problem Formulation}
Given an event sequence $\mathcal{D}_n = \{(t_i, e_i)\}_{i=1}^n$ where $t_i$ denotes timestamp and $e_i \in \{1,\ldots,E\}$ represents event type, we aim to model the conditional intensity function $\lambda^e(t|\mathcal{H}_t)$ while learning dynamic dependencies between events. Let $G_t = (V, E_t)$ represent the time-varying graph structure where $V$ are event types and $E_t$ captures evolving dependencies at time $t$. The joint optimization objective is:

\begin{equation}
\min_{\theta, G_t} \mathcal{L}(\theta) = -\sum_{i=1}^n \log \lambda^{e_i}(t_i|\mathcal{H}_{t_i}) + \int_0^T \sum_{e=1}^E \lambda^e(t|\mathcal{H}_t)dt + R(G_t)
\label{eq:objective}
\end{equation}

where $\theta$ are model parameters and $R(G_t)$ is a sparsity-inducing regularizer. Our key insight is leveraging spiking neural dynamics to naturally model these temporal dependencies - membrane potentials of LIF neurons serve as adaptive basis functions that can efficiently capture evolving relationships between events.

\subsection{Learning Algorithm}
Our algorithm comprises three coupled components that jointly optimize the objective in Eq.~\ref{eq:objective}:

\textbf{Membrane Potential Dynamics.} The membrane potential $v_i(t)$ of neuron $i$ evolves as:
\begin{equation}
\tau_m \frac{dv_i(t)}{dt} = -\alpha(v_i(t) - v_{\text{rest}}) + I_i^{\text{syn}}(t) + I_i^{\text{ext}}(t)
\label{eq:membrane}
\end{equation}
where $\tau_m$ is the membrane time constant and $I_i^{\text{syn}}(t)$ encodes temporal dependencies through:
\begin{equation}
I_i^{\text{syn}}(t) = \sum_{j \in \text{pre}(i)} w_{ij}(t) \sum_{k} \epsilon(t - t_j^k) \exp(-\beta(t - t_j^k))
\label{eq:synaptic}
\end{equation}
Here $w_{ij}(t)$ represents synaptic weights and $t_j^k$ are spike times.

\textbf{STDP Learning.} Synaptic weights adapt via a regularized STDP rule:
\begin{equation}
\Delta w_{ij} = \eta \cdot \text{STDP}(\Delta t_{ij}) \cdot \left(1 - \frac{w_{ij}}{w_{\text{max}}}\right)^\alpha \exp(-\beta|w_{ij}|)
\label{eq:stdp}
\end{equation}
where $\Delta t_{ij}$ is the spike timing difference and $\eta$ controls learning rate.

\textbf{Graph Structure Estimation.} Edge probabilities between neuron pairs $(i,j)$ are computed as:
\begin{equation}
p_{ij}(t) = \text{Softmax}\left(\frac{1}{T}\sum_{k=1}^K \phi(s_i^k, s_j^k)\right), \end{equation}
\begin{equation}
\phi(s_i, s_j) = \sum_{t_i \in s_i} \sum_{t_j \in s_j} \exp\left(-\frac{|t_i - t_j|}{\tau}\right)
\label{eq:graph}
\end{equation}
where $s_i^k, s_j^k$ are spike trains and $\tau$ controls the temporal kernel width.

\subsection{Intensity Function Estimation}
We model the conditional intensity function as a combination of structural and temporal components:

\begin{equation}
\lambda^v(t|\mathcal{H}_t) = f\left(\sum_{u \in \mathcal{N}(v)} \alpha_{uv}(t) h_u(t) + \beta_v \delta(t-t_n)\right)
\label{eq:intensity}
\end{equation}

where $h_u(t)$ are neural embeddings derived from membrane potentials, $\alpha_{uv}(t)$ are learned attention weights capturing dynamic node interactions, and $\beta_v$ models type-specific effects. The non-linear activation $f(\cdot)$ ensures positive intensity values.

\textbf{Theorem 1.} \textit{Under regularity conditions (Supplementary A.3), with probability 1, the algorithm converges to a local optimum of the objective in Eq.~\ref{eq:objective}.}

The proof leverages stochastic approximation theory and properties of our regularized STDP rule to establish almost sure convergence.

\subsection{Implementation}
For stable and efficient training, we introduce three key optimizations:

1) Adaptive time-stepping that bounds membrane potential changes:
\begin{equation}
\Delta t = \min\left(\frac{\tau_{\text{min}}}{\max_i |v_i(t)|}, \Delta t_{\text{max}}\right)
\label{eq:timestep}
\end{equation}

2) Differentiable spike approximation using a surrogate gradient:
\begin{equation}
\frac{\partial s}{\partial v} \approx \sigma'(v-v_{\text{th}}) \cdot \text{clip}(\alpha|v-v_{\text{th}}|, 0, 1)
\label{eq:surrogate}
\end{equation}

3) Event-driven updates with priority queues for $O(\log N)$ complexity per spike.

The complete algorithm, including initialization and hyperparameter settings, is detailed in Algorithm \ref{alog:sdgn}.

\begin{algorithm}[t]
\caption{SDGN Training}
\label{alg:sdgn}
\begin{algorithmic}[1]
\Require Event sequence $\mathcal{D}_n$, learning rate $\eta$, time constants $\tau_m, \tau_{\text{STDP}}$
\Ensure Trained model parameters $\theta$, graph structure $G_t$
\State Initialize: membrane potentials $v_i=v_{\text{rest}}$, weights $w_{ij}\sim \mathcal{U}(-0.1,0.1)$
\For{each training batch}
    \State $t \gets 0$
    \While{$t < T$}
        \State Compute $\Delta t$ using Eq.~\ref{eq:timestep}
        \State Update membrane potentials using Eq.~\ref{eq:membrane}
        \For{each spiking neuron $i$}
            \State Generate spike: $s_i(t) \gets 1$
            \State Reset: $v_i(t) \gets v_{\text{reset}}$
            \For{each presynaptic neuron $j$}
                \State Update $w_{ij}$ using Eq.~\ref{eq:stdp}
                \State Update edge probability $p_{ij}$ using Eq.~\ref{eq:graph}
            \EndFor
        \EndFor
        \State Update intensity $\lambda^v(t|\mathcal{H}_t)$ using Eq.~\ref{eq:intensity}
        \State $t \gets t + \Delta t$
    \EndWhile
    \State Update model parameters $\theta$ using gradient descent
\EndFor
\State \Return $\theta$, $G_t$
\end{algorithmic}
\end{algorithm}

\subsection{Dynamic Neural Basis Construction}
We propose representing temporal dependencies through adaptive neural bases that can capture the underlying low-dimensional manifold structure. Let $\Phi_t: \mathbb{R}^+ \to \mathbb{R}^d$ be our time-varying basis representation:

\begin{equation}
\Phi_t(s) = \sum_{i=1}^N h_i(t) \kappa(s - t_i)
\label{eq:basis}
\end{equation}

where $h_i(t)$ are STDP-learned coefficients, $\kappa(s)$ is the membrane response kernel, and $t_i$ are spike times. The basis functions adapt to temporal patterns through membrane dynamics:

\begin{equation}
\tau_m \frac{dv_i(t)}{dt} = -\alpha(v_i(t) - v_{\text{rest}}) + \sum_{j \in \text{pre}(i)} w_{ij}(t) \sum_{k} \epsilon(t - t_j^k) \exp(-\beta(t - t_j^k))
\label{eq:membrane_dynamics}
\end{equation}

This enables precise temporal sensitivity through exponential decay terms and adaptive connectivity through $w_{ij}(t)$.

\subsection{Graph Structure Learning}
We model temporal dependencies through a framework that combines functional connectivity with spike timing relationships. For event types $(i,j)$, we define their dependency strength:

\begin{equation}
\mathcal{S}_{ij}(t) = \mathbb{E}\left[\int_0^T K(t-s)\text{d}N_i(s) \int_0^T K(t-u)\text{d}N_j(u)\right]
\label{eq:dependency}
\end{equation}

where $N_i(t)$ is the counting process for event type $i$ and $K(t)$ is learned from spike responses. The dynamic graph structure is then estimated through thresholding $\mathcal{S}_{ij}(t)$, allowing the model to adapt to evolving temporal relationships between events.

\subsection{Learning Algorithm}
Our model learns through three coupled mechanisms:
\begin{equation}
\Delta w_{ij} = \eta \cdot \text{STDP}(\Delta t_{ij}) \cdot \left(1 - \frac{w_{ij}}{w_{\text{max}}}\right)^\alpha \exp(-\beta|w_{ij}|)
\label{eq:weight_update}
\end{equation}

\begin{equation}
p_{ij}(t) = \text{Softmax}\left(\frac{1}{T}\sum_{k=1}^K \sum_{t_i \in s_i^k} \sum_{t_j \in s_j^k} \exp\left(-\frac{|t_i - t_j|}{\tau}\right)\right)
\label{eq:edge_prob}
\end{equation}

\begin{equation}
\lambda^v(t|\mathcal{H}_t) = f\left(\sum_{u \in \mathcal{N}(v)} \alpha_{uv}(t) h_u(t) + \beta_v \delta(t-t_n)\right)
\label{eq:intensity}
\end{equation}

where $\Delta t_{ij}$ is the spike timing difference, $s_i^k$ are spike trains, $h_u(t)$ are neural embeddings, and $\alpha_{uv}(t), \beta_v$ are learnable parameters. The weight regularization in Eq.~\ref{eq:weight_update} prevents weight explosion while maintaining sparse connectivity.

\subsection{Optimization}
We employ three techniques for stable training:

1) Adaptive time discretization that bounds potential changes:
\begin{equation}
\Delta t = \min\left(\frac{\tau_{\text{min}}}{\max_i |v_i(t)|}, \Delta t_{\text{max}}\right)
\label{eq:timestep}
\end{equation}

2) Differentiable spike approximation:
\begin{equation}
\frac{\partial s}{\partial v} = \sigma'(v-v_{\text{th}}) \cdot \text{clip}(\alpha|v-v_{\text{th}}|, 0, 1)
\label{eq:spike_grad}
\end{equation}

3) Priority queue-based spike propagation achieving $O(\log N)$ complexity per spike, enabling efficient scaling to large networks.

\section{Theoretical Analysis}

\textbf{Theorem 1} (Learning Capacity). \textit{Consider a SDGN with $N$ neurons where each neuron has maximum fan-out $k = \Omega(log N)$. Under random sparse connectivity, the network can learn $m = O(N^2)$ distinct temporal patterns with probability at least $1 - exp(-N)$, where each pattern consists of a sequence of spike times in $[0,T]$.}

\begin{proof}
Let $\mathcal{P} = \{p_1,...,p_m\}$ be a set of m distinct temporal patterns where each $p_i$ represents a finite sequence of spike times in $[0,T]$. First, we characterize the membrane potential $v_j^i(t)$ of neuron $j$ in response to pattern $p_i$:
\begin{equation}
v_j^i(t) = \sum_{s \in p_i} w_{js} \exp\left(-\frac{t-s}{\tau_m}\right)H(t-s) + v_{\text{rest}}
\end{equation}
where $H(t)$ is the Heaviside function and weights $w_{js} \sim \mathcal{U}(-\delta,\delta)$ for connected neurons (otherwise 0). We define the activation matrix $A \in \mathbb{R}^{m×N}$ where $A_{ij}$ represents the total activation:
\begin{align}
A_{ij} &= \int_0^T v_j^i(t) dt \\
&= \sum_{s \in p_i} w_{js} \tau_m\left(1-\exp\left(-\frac{T-s}{\tau_m}\right)\right) + v_{\text{rest}}T
\end{align}

 For matrix $A$ to encode $m$ patterns, it must have rank $m$. Consider the $k$-sparse weight matrix $W$ where each column has $k$ non-zero entries. By standard results in random matrix theory [1], for $k = \Omega(log N)$:
\begin{equation}
P(\sigma_{\text{min}}(A) \ge \epsilon) \ge 1 - \exp(-ck)
\end{equation}
where $\sigma_{\text{min}}(A)$ is the minimum singular value and $c > 0$ is a constant. The number of learnable patterns m is bounded by the rank of $A$. For $k$-sparse connectivity:
\begin{equation}
\text{rank}(A) \le \min(m, kN)
\end{equation}

Therefore, with $k = \Omega(log N)$, the network can learn $m = O(N^2)$ patterns with probability at least $1 - exp(-N)$.
\end{proof}

\textbf{Theorem 2} (Temporal Resolution). \textit{For a spiking neural network with membrane time constant $\tau_m$ and decay rate $\alpha$, the minimum detectable time difference between spikes is bounded by:}
\begin{equation}
|\Delta t| \geq \frac{\tau_m}{\alpha \max_i |v_i(t)|} \epsilon
\label{eq:temporal_bound}
\end{equation}
\textit{where $\epsilon$ is the detection threshold.}

\begin{proof}
Consider two spikes at times $t_1$ and $t_2 = t_1 + \Delta t$. For reliable detection, their membrane potentials must differ by at least $\epsilon$:
\begin{equation}
|v(t_2) - v(t_1)| \geq \epsilon
\label{eq:detection}
\end{equation}

From the membrane dynamics equation:
\begin{equation}
\tau_m \frac{dv(t)}{dt} = -\alpha(v(t) - v_{\text{rest}}) + I(t)
\label{eq:membrane_dynamics}
\end{equation}

Assuming negligible input current between spikes, the solution is:
\begin{equation}
v(t_2) = v_{\text{rest}} + (v(t_1) - v_{\text{rest}})e^{-\alpha \Delta t/\tau_m}
\label{eq:solution}
\end{equation}

The potential difference is thus:
\begin{equation}
|v(t_2) - v(t_1)| = |v(t_1) - v_{\text{rest}}|(1 - e^{-\alpha \Delta t/\tau_m})
\label{eq:difference}
\end{equation}

From Eq.~\ref{eq:detection} and noting $|v(t_1) - v_{\text{rest}}| \leq \max_i |v_i(t)|$:
\begin{equation}
\max_i |v_i(t)|(1 - e^{-\alpha \Delta t/\tau_m}) \geq \epsilon
\label{eq:inequality}
\end{equation}

For small $\Delta t$, using Taylor expansion:
\begin{equation}
1 - e^{-\alpha \Delta t/\tau_m} \approx \frac{\alpha \Delta t}{\tau_m}
\label{eq:taylor}
\end{equation}

Substituting and solving for $\Delta t$ yields the bound in Eq.~\ref{eq:temporal_bound}.
\end{proof}

\textbf{Theorem 3 (Convergence Rate).} \textit{Under the STDP learning rule with regularization, assuming bounded martingale noise and Lipschitz continuous regularization, the synaptic weights converge to a local optimum at rate $O(1/\sqrt{T})$ where T is the number of observed spikes.}

\begin{proof}
The weight updates follow the stochastic approximation:

\begin{equation}
w^{t+1} = w^t + \eta_t[H(w^t) + M^t]
\end{equation}

where $H(w) = \mathbb{E}[\text{STDP}(\Delta t)\text{reg}(w)]$, $M^t$ is a martingale difference sequence with $\mathbb{E}[M^t|\mathcal{F}_t] = 0$ and bounded variance $\sigma^2$.

The regularization term satisfies Lipschitz continuity:
\begin{equation}
\|H(w_1) - H(w_2)\| \leq L\|w_1 - w_2\|
\end{equation}

For Lyapunov function $V_t = \|w^t - w^*\|^2$, expanding and taking conditional expectation:
\begin{equation}
\mathbb{E}[V_{t+1}|\mathcal{F}_t] = V_t + 2\eta_t\langle w^t - w^*, H(w^t)\rangle + \eta_t^2\mathbb{E}[\|H(w^t) + M^t\|^2|\mathcal{F}_t]
\end{equation}

By strong monotonicity and bounded noise:
\begin{equation}
\mathbb{E}[V_{t+1}|\mathcal{F}_t] \leq (1 - \mu\eta_t)V_t + 2\sigma^2\eta_t^2
\end{equation}

Using learning rate $\eta_t = 1/\sqrt{t}$ and applying stochastic approximation theory:
\begin{equation}
\mathbb{E}[V_T] \leq \frac{C}{\sqrt{T}}
\end{equation}

where $C$ depends on $\mu$, $L$ and $\sigma^2$. Therefore:
\begin{equation}
\mathbb{E}[\|w^T - w^*\|^2] = O(1/\sqrt{T})
\end{equation}

This rate is optimal in the presence of martingale noise, matching lower bounds for stochastic optimization.
\end{proof}

\section{Experiments and Empirical Results}

\subsection{Datasets}

\textbf{Synthetic Datasets:}  To systematically assess the performance of our graph estimation algorithm, we design synthetic networks that reflect diverse neuronal connectivity patterns. We created a synthetic dataset to simulate multivariate event streams using dynamic random graphs. Each node in the graph generates event streams through a Hawkes process, with the intensity function dependent on its connectivity. The graph's structure evolves over time, incorporating random temporal edges from the previous time step.

\textbf{Real-world Datasets:}  We use the proposed model to analyze multivariate event sequence data across various domains. The datasets include NYC TAXI, Reddit, Earthquake, Stack Overflow, and 911 Calls.

\subsection{Baselines} To assess the performance of the proposed SDGNN, we compare it with three types of baseline models: 

\begin{itemize}
    \item \textbf{Traditional Statistical Models} encompasses stochastic models that use parametric intensity functions, including the Poisson Process \cite{poisson}, Hawkes Process \cite{hawkes}, and self-correcting processes \cite{isham1979self}.
    \item \textbf{Neural Temporal Point Process Models} includes models such as RMTPP \cite{du2016recurrent} and NHP \cite{mei2017}, which leverage a simple RNN to capture the historical event sequences, and THP \cite{zuo2020transformer}, which utilizes self-attention for sequential event embedding.
    \item \textbf{Neural Graphical Event Models} which include GRTPP \cite{yoon2023} and SDGNN.
\end{itemize}


%

\begin{table*}[]
\centering
\caption{Performance Comparison of Various Temporal Point Process Models on Real-World Datasets}
\label{tab:real-world}
\resizebox{0.95\textwidth}{!}{%
\begin{tabular}{l l c c c c c}
\hline
\textbf{Model} & \textbf{Description} & \textbf{NYC Taxi} & \textbf{Reddit} & \textbf{\begin{tabular}[c]{@{}c@{}}Stack\\ Overflow\end{tabular}} & \textbf{Earthquake} & \textbf{911} \\ \hline
\textit{\textbf{Poisson}} & \begin{tabular}[c]{@{}l@{}}Classical model,\\ assumes independent events\end{tabular} & $1.051$ & $307.4$ & $11.91$ & $103.7$ & $5.488$ \\ \hline
\textit{\textbf{Hawkes}} & \begin{tabular}[c]{@{}l@{}}Classical model,\\ self-exciting process\end{tabular} & $1.021$ & $48.54$ & $12.32$ & $13.24$ & $5.822$ \\ \hline
\textit{\textbf{Self-Correcting}} & \begin{tabular}[c]{@{}l@{}}Classical model,\\ self-correcting process\end{tabular} & $1.042$ & $62.42$ & $11.35$ & $11.24$ & $5.543$ \\ \hline
\textit{\textbf{RMTPP}} \cite{du2016recurrent} & \begin{tabular}[c]{@{}l@{}}Uses RNN to model\\ event timings and markers\end{tabular} & 
\begin{tabular}[c]{@{}c@{}}$0.947$\\$\pm 0.003$\end{tabular} & 
\begin{tabular}[c]{@{}c@{}}$16.18$\\$\pm 0.224$\end{tabular} & 
\begin{tabular}[c]{@{}c@{}}$9.782$\\$\pm 0.052$\end{tabular} & 
\begin{tabular}[c]{@{}c@{}}$9.105$\\$\pm 0.19$\end{tabular} & 
\begin{tabular}[c]{@{}c@{}}$6.231$\\$\pm 0.052$\end{tabular} \\ \hline
\textit{\textbf{NHP}} \cite{mei2017neural} & \begin{tabular}[c]{@{}l@{}}Continuous-time LSTM\\ for self-modulating\\ multivariate processes\end{tabular} & 
\begin{tabular}[c]{@{}c@{}}$0.932$\\$\pm 0.003$\end{tabular} & 
\begin{tabular}[c]{@{}c@{}}$15.88$\\$\pm 0.182$\end{tabular} & 
\begin{tabular}[c]{@{}c@{}}$9.832$\\$\pm 0.082$\end{tabular} & 
\begin{tabular}[c]{@{}c@{}}$9.081$\\$\pm 0.033$\end{tabular} & 
\begin{tabular}[c]{@{}c@{}}$5.449$\\$\pm 0.012$\end{tabular} \\ \hline
\textit{\textbf{THP}} \cite{zuo2020transformer} & \begin{tabular}[c]{@{}l@{}}Transformer-based model\\ for long-term dependencies\end{tabular} & 
\begin{tabular}[c]{@{}c@{}}$0.842$\\$\pm 0.005$\end{tabular} & 
\begin{tabular}[c]{@{}c@{}}$16.56$\\$\pm 0.309$\end{tabular} & 
\begin{tabular}[c]{@{}c@{}}$4.99$\\$\pm 0.019$\end{tabular} & 
\begin{tabular}[c]{@{}c@{}}$8.988$\\$\pm 0.107$\end{tabular} & 
\begin{tabular}[c]{@{}c@{}}$5.329$\\$\pm 0.015$\end{tabular} \\ \hline
\textit{\textbf{GRTPP}} \cite{yoon2023} & \begin{tabular}[c]{@{}l@{}}Encodes sequences\\ into graph structures\end{tabular} & 
\begin{tabular}[c]{@{}c@{}}$0.599$\\$\pm 0.021$\end{tabular} & 
\begin{tabular}[c]{@{}c@{}}$11.87$\\$\pm 0.152$\end{tabular} & 
\begin{tabular}[c]{@{}c@{}}$5.522$\\$\pm 0.032$\end{tabular} & 
\begin{tabular}[c]{@{}c@{}}$7.415$\\$\pm 0.036$\end{tabular} & 
\begin{tabular}[c]{@{}c@{}}$5.159$\\$\pm 0.033$\end{tabular} \\ \hline
\rowcolor[HTML]{FFFC9E} 
\textit{\textbf{SDGN (Ours)}} & \begin{tabular}[c]{@{}l@{}}Dynamic graph estimation\\ using SNNs\end{tabular} & 
\textbf{\begin{tabular}[c]{@{}c@{}}$0.422$\\$\pm 0.022$\end{tabular}} & 
\textbf{\begin{tabular}[c]{@{}c@{}}$9.13$\\$\pm 0.106$\end{tabular}} & 
\textbf{\begin{tabular}[c]{@{}c@{}}$4.106$\\$\pm 0.055$\end{tabular}} & 
\textbf{\begin{tabular}[c]{@{}c@{}}$6.351$\\$\pm 0.028$\end{tabular}} & 
\textbf{\begin{tabular}[c]{@{}c@{}}$4.033$\\$\pm 0.025$\end{tabular}} \\ \hline
\end{tabular}%
}
\end{table*}

\subsection{Synthetic Data: Performance in Estimating Graph}
\begin{figure*}
    \centering
    \includegraphics[width=0.82\textwidth]{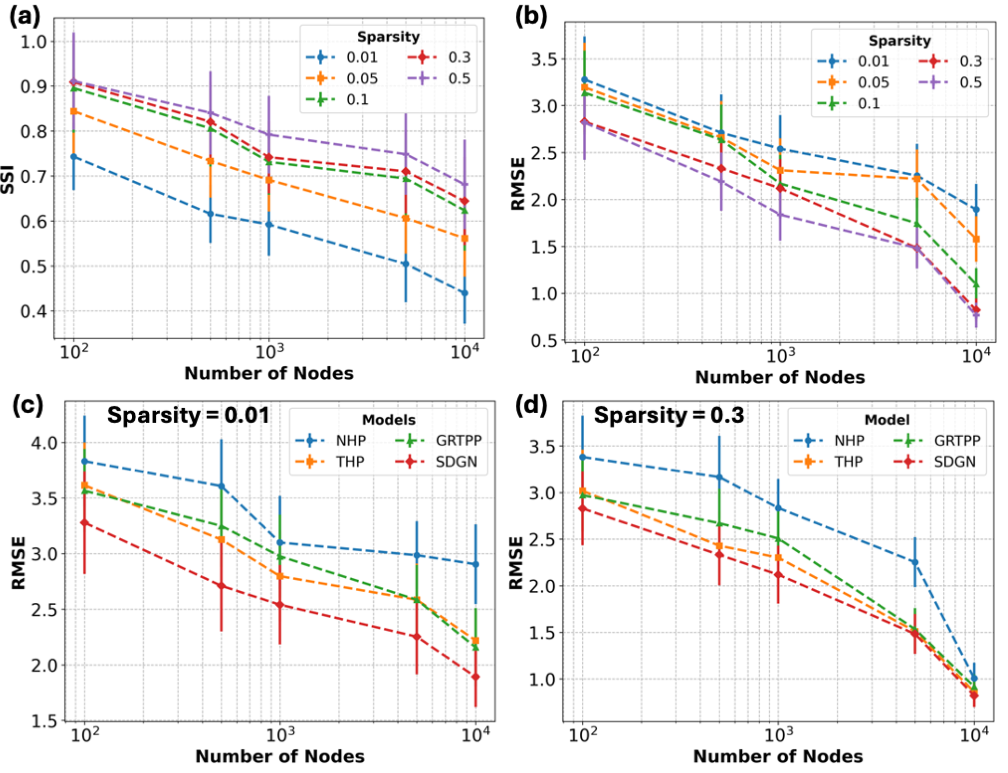}
    \caption{Performance of (a) the graph estimation algorithm using Structural Similarity Index (SSI) between the underlying and estimated temporal graphs vs the Number of Nodes (b) RMSE for the event prediction task for different sparsity levels for the synthetic dataset (c), (d) Performance of Event Prediction with other baselines on the synthetic dataset for sparsity = 0.01, 0.3 respectively}
    \label{fig:ssi}
\end{figure*}

We evaluate the SDGN model's ability to estimate the underlying graphical structure using synthetic data, measured by the Structural Similarity Index (SSI). The SSI compares the adjacency matrices of the true graph \( G \) and the estimated graph \( G' \), incorporating local means, variances, and covariances. SSI ranges from -1 to 1, with 1 indicating perfect similarity and -1 complete dissimilarity. Figure \ref{fig:ssi}(a) shows SSI for varying sparsity levels and increasing node counts. SSI generally decreases with more nodes, indicating lower structural similarity in larger networks. Higher sparsity results in lower SSI values, suggesting less similarity in sparser networks, while denser networks maintain higher SSI values, reflecting greater similarity. This trend is more pronounced in smaller networks. Error bars indicate SSI variability, with greater variances in sparser networks. Figure \ref{fig:ssi}(b) depicts model performance on event prediction with the synthetic dataset. Estimating the graph accurately is more challenging for sparser, larger networks, resulting in decreased model performance. Conversely, a lower Root Mean Square Error (RMSE) is observed where SSI is high, indicating an inverse correlation between SSI and RMSE. Thus, better-estimated models correlate with improved event prediction performance.

\textbf{Performance Comparison with Other Baselines:} We also evaluated the performance of the Spiking Dynamic Graph Network (SDGN) against other baseline models on the synthetic dataset across varying levels of sparsity and different numbers of nodes. The results are depicted in Figures \ref{fig:ssi}(c) and \ref{fig:ssi}(d). Our findings indicate that for sparser graphs, SDGN significantly outperforms the baseline models. However, as the graph density increases, the performance difference diminishes, suggesting that the graphical structure may not provide substantial benefits when the underlying graph is large and fully connected.

\begin{figure*}
    \centering
    \includegraphics[width=0.85\textwidth]{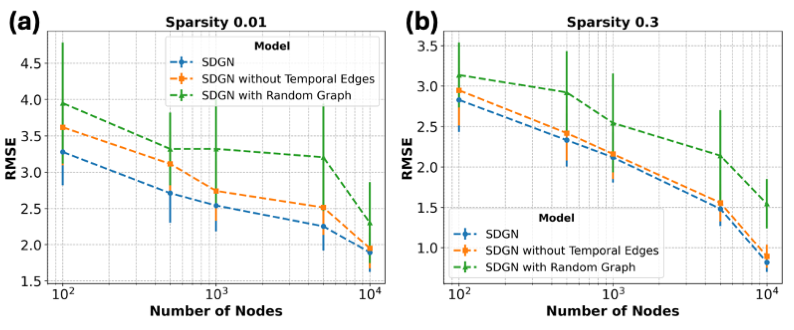}
    \caption{Performance of the event prediction vs Number of Nodes for different sparsity levels for the synthetic dataset. The results are compared across models with random graphs and only spatial functional graphs}
    \label{fig:perf_abl}
\end{figure*}

\subsection{Ablation Study: Event Prediction Performance with Synthetic Dataset}

Here we compare the performance of the graph estimation algorithm of the SDGN method with estimating just the spatial functional graphs and with random graphs. We evaluate the performance on the synthetic dataset with varying sparsity levels and an increasingly large number of nodes. 
Figure \ref{fig:perf_abl} presents the RMSE as a function of the number of nodes for various models and sparsity levels in a synthetic dataset. The subfigures (a) through (d) correspond to different sparsity levels. Each subfigure compares the performance of three models: SDGN, SDGN without Temporal Edges, and SDGN with Random Graph. RMSE decreases as the number of nodes increases for all sparsity levels, indicating improved prediction accuracy in larger networks. Also, the SDGN model consistently outperforms the other models, exhibiting lower RMSE values. These results highlight the effectiveness of the SDGN model in leveraging temporal edges and structural information to improve prediction accuracy, with more pronounced benefits observed as network size increases. The error bars illustrate the variability in RMSE measurements, showing larger variances for models with random graphs and at higher sparsity levels.

\subsection{Performance on Real-world Datasets}

In this section, we compare the performance of different temporal point process models on several real-world datasets: NYC Taxi, Reddit, Stack Overflow, Earthquake, and 911. The models evaluated include Poisson, Hawkes, Self-correcting, RMTPP, NHP, THP, GRTPPA, and our proposed SDGN method. The results are shown in Table \ref{tab:real-world}. The results demonstrate that SDGN outperforms the other models across all datasets, achieving the lowest error rates. These results highlight the superior performance of our model in effectively modeling and predicting real-world temporal events.

\section{Conclusions}

In this paper, we presented the Spiking Dynamic Graph Network (SDGN), a novel approach for modeling and predicting TPPs. SDGN leverages the event processing capabilities of RSNNs and the online learning abilities of STDP to dynamically estimate temporal functional graphs, capturing complex interdependencies among event streams. Our evaluations on synthetic and real-world datasets demonstrate SDGN's superior predictive accuracy over state-of-the-art methods.

\textbf{Limitations and Future work:} The dynamic estimation of graph structures and stability issues of STDP introduce complexity and computational overhead, limiting the scalability of SDGN for very large graphs. Unlike DNNs, SDGN's reliance on precise spike timing and dynamic graph updates poses specific challenges in computational complexity and the need for robust handling of timing inaccuracies. Furthermore, as indicated by our results, SDGN does not demonstrate a significant performance advantage for dense graphs, where the benefits of dynamic graph estimation are less pronounced. This highlights the need to study the scalability of SDGN and explore the potential parallelization of the algorithm. Future work could also investigate the effect of higher-order interactions and how they can be leveraged to efficiently estimate functional neighborhoods, providing an interesting direction for enhancing the model's capabilities.

\bibliography{ref}
\bibliographystyle{unsrt}

\newpage
\appendix

\section{Appendix}

\section{Detailed Methods}
\label{sec:method}


\begin{itemize}
    \item \textbf{Functional Graphical Model: } We consider each event stream (observation) is modeled as a realization of a multivariate Gaussian Process (MGP)
    \item The goal is to estimate the conditional independence structure among multivariate random functions, which can be interpreted as nodes in a graph, with edges representing conditional dependencies.
    \item \textbf{Neighborhood Selection Approach: }We use a neighborhood selection method inspired by the work of Meinshausen and Bühlmann (2006) for vector-valued Gaussian graphical models.
    \item This approach estimates the neighborhood of each node (i.e., the set of adjacent nodes) through sparse regression.
    \item In this work we are modeling the event streams as functional data. For functional data, this translates into a function-on-function regression problem, approximated by projecting the functional data onto a finite-dimensional basis and solving a vector-on-vector regression with a group lasso penalty.
    \item \textbf{RSNN + STDP:} We use a RSNN to project the infinite dimensional functional data onto a finite dimensional subspace 
    \item The use of STDP can be used to get the temporal connectivity of the model. It can also be later on leveraged for non-stationary input signals 
    \item This projection helps us transform the function-on-function regression into a vector-on-vector regression problem.
    \item \textbf{Key contribution - using RSNN+ STDP as a functional basis for dimension reduction: } 
\end{itemize}


\subsection{Problem Definition}
We define an event stream as \(\mathcal{D}_n = \{(t_i, e_i)\}_{i=1}^n\), where \(t_i\) is the timestamp, \(e_i \in \{1, ..., E\}\) is the event type, and \(E\) is the total number of event types. Each event is a pair \((t_i, e_i)\). The challenge is that we do not know the correlations between these event types, meaning we do not know how the occurrence of certain event types affects the probability of other types occurring.

The problem is to observe the sequence of events and predict the timing and type of the next events. Specifically, we aim to estimate the rate at which events of each type occur over time and use this information to make predictions about future events. This involves developing a model that can learn from the event history \(\mathcal{H}_t = \{(t_i, e_i)|t_i < t\}\) and accurately forecast the next events in the stream.

Our approach, termed the Spiking Dynamic Graph Network (SDGN), comprises three main components:

\begin{itemize}
    \item \textbf{Constructing Dynamic Temporal Functional Graphs:} We build dynamic temporal functional graphs from multivariate event sequences, capturing both temporal and relational dependencies among events.
    \item \textbf{Learning Dynamic Node Embeddings:} Using SNNs, we learn dynamic node embeddings that encode both temporal and relational information, enhancing the representation of event sequences.
    \item \textbf{Estimating Intensity Functions for Future Events:} Based on the learned embeddings, we estimate the intensity functions for predicting future events.
\end{itemize}

\subsection{Background}

\subsubsection{Functional Graphical Models}

Consider a closed interval $\mathcal{T} \subseteq \mathbb{R}$ and a Hilbert subspace $\mathbb{H}$ of $\mathcal{L}^2(\mathcal{T})$. Since $\mathcal{L}^2(\mathcal{T})$ is a separable Hilbert space, $\mathbb{H}$ and $\mathbb{H}^n$ are also separable for any finite \( n \). Let $(\Omega, \mathcal{F}, \mathbb{P})$ be a probability space and $\bm{g}:\Omega \mapsto \mathbb{H}^{p}$ a Gaussian random element. For any $\bm{h} \in \mathbb{H}$, $\langle \bm{g}, \bm{h} \rangle$ is a Gaussian random variable. We can write $\bm{g}$ as $\bm{g}(\omega,t) = (g_1(\omega,t), \ldots, g_p(\omega,t))$, where $(\omega,t) \in \Omega \times \mathcal{T}$, and for all $\omega \in \Omega$ and \( j \in [p] \), \( g_j(\omega, \cdot) \in \mathbb{H} \). We simplify the notation by writing $\bm{g}(t) = (g_1(t), \ldots, g_p(t))$ for $t \in \mathcal{T}$, and sometimes just $\bm{g} = (g_1, \ldots, g_p)$.

Assuming $\bm{g}$ has a zero mean, i.e., $\mathbb{E}[g_j] = 0$ for all $j \in [p]$, and given the Gaussian property, $\mathbb{E}[\Vert \bm{g} \Vert^2] < \infty$. Thus, for each \( j \in [p] \), we define the covariance operator of \( g_j \) as:
\[ 
\mathscr{K}_j \coloneqq \mathbb{E} [ g_j \otimes g_j ], 
\]
with $\mathscr{K}_j \in \mathcal{B}_{\text{HS}}(\mathbb{H})$. For any index sets \( I, I_1, I_2 \subseteq [n] \), define:
\[
\mathscr{K}_I \coloneqq \mathbb{E} \left[ \left( g_j \right)_{j \in I} \otimes \left( g_j \right)_{j \in I} \right], \quad \mathscr{K}_{I_1, I_2} \coloneqq \mathbb{E} \left[ \left( g_j \right)_{j \in I_1} \otimes \left( g_j \right)_{j \in I_2} \right].
\]

Following \cite{Qiao2015Functional}, we define the conditional cross-covariance function as:
\[
C_{jl}(t^{\prime}, t) = \text{Cov} \left( g_j(t^{\prime}), g_l(t) \mid g_k(\cdot), k \neq j, l \right).
\]

Let \( G = (V, E) \) be an undirected graph where \( V = [p] \) represents the nodes and \( E \subseteq V^2 \) the edges. The edge set \( E \) encodes the pairwise Markov property of $\bm{g}$ \cite{Lauritzen1996Graphical} if:
\[
E = \{ (j, l) \in V^2 : j \neq l \text{ and } g_j \not \independent g_l \mid \{g_k\}_{k \neq j, l} \}.
\]

Given \( n \) i.i.d. random copies \( \{\bm{g_i}(\cdot)\}_{i=1}^n \) of $\bm{g}$, our goal is to estimate the edge set $E$. \cite{Qiao2015Functional} proposed a functional graphical lasso procedure for this estimation. We propose a neighborhood selection approach, defining the neighborhood of node \( j \) as:
\[
\mathscr{N}_j \coloneqq \{k : (j,k) \in E\}.
\]

\subsubsection{Receptive Fields in Neural Networks}

In the context of neural networks, especially spiking neural networks (SNNs), a receptive field refers to the specific region of sensory input that a particular neuron responds to. In other words, it is the subset of input features or input space that influences the activity of a given neuron.

For example, in a visual system, the receptive field of a neuron in the visual cortex would correspond to a specific area of the visual field that, when stimulated, affects the neuron's firing rate. Similarly, in an SNN, the receptive field of a neuron can be thought of as the set of input spikes or the spatiotemporal patterns of input that the neuron is sensitive to. In the algorithm described for estimating the spatial and temporal functional graphs using a Recurrent Spiking Neural Network (RSNN), the receptive field can be interpreted as follows:

\begin{itemize}
    \item In an RSNN, each neuron receives inputs from a certain subset of the input space (parallel event streams). The neuron's response to these inputs over time constitutes its receptive field.
    \item The receptive field is characterized by the spiking activity of the neuron in response to its inputs, capturing the spatiotemporal dynamics of how the neuron processes the input data.
    \item To calculate the receptive field for each neuron, we analyze the spike trains generated by the neuron in response to the input event streams.
    \item The receptive field \(f_i(x)\) can be represented as the spiking response of neuron \(i\) to the input feature \(x\). This involves tracking how the neuron's spiking activity correlates with specific patterns in the input data.
\end{itemize}

\subsection{Graph Estimation}

\subsubsection{Spatial Functional Neighborhood Estimation}

To estimate the functional graphical model, we adopt a neighborhood selection procedure inspired by \cite{Besag1975Statistical} and extend it to functional data. For each node \(j \in [p]\), we aim to identify the set of nodes \(k \neq j\) such that there is a direct dependency between \(g_j\) and \(g_k\), conditioned on all other nodes.

Consider the conditional expectation \(\mathbb{E}[g_j \mid \bm{g}_{-j}]\) for \(j \in [p]\). By the Doob-Dynkin representation theorem, there exists a measurable map \(\mathscr{B}_j : \mathbb{H}^{p-1} \to \mathbb{H}\) such that
\[
\mathbb{E}[g_j \mid \bm{g}_{-j}] = \mathscr{B}_j(\bm{g}_{-j}) \quad \text{almost surely}.
\]
Under Gaussianity, \(\mathscr{B}_j \in \mathcal{B}(\mathbb{H}^{p-1}, \mathbb{H})\), and \(e_j = g_j - \mathbb{E}[g_j \mid \bm{g}_{-j}]\) is independent of \(\bm{g}_{-j}\).

Assume that \(\mathscr{B}_j\) belongs to the class of Hilbert-Schmidt operators \(\mathcal{B}_{\text{HS}}(\mathbb{H}^{p-1}, \mathbb{H})\), ensuring a finite-dimensional approximation. This leads to the representation:
\[
g_j(t) = \sum_{k \neq j} \int_{\mathcal{T}} \beta_{jk}(t, t') g_k(t') \, dt' + e_j(t),
\]
where \(e_j(t) \independent g_k(t)\) for \(k \neq j\), and \(\Vert \beta_{jk}(t, t') \Vert_{\text{HS}} < \infty\). Thus, the neighborhood of node \(j\) is defined as:
\[
\mathscr{N}_j = \{k \in [p] \backslash \{j\} : \Vert \beta_{jk} \Vert_{\text{HS}} > 0\}.
\]

To estimate each neighborhood, we regress \(g_j\) on \(\{ g_k : k \in [p] \backslash \{j\} \}\) using a penalized functional regression approach. Specifically, we solve the following optimization problem:
\begin{align}
    \bm{\hat{B}}_j &= \arg \min_{\bm{B}_j} \left\{ \frac{1}{2n} \sum_{i=1}^n \left\Vert g_{ij} - \sum_{k \neq j} \int_{\mathcal{T}} \beta_{jk}(t, t') g_{ik}(t') \, dt' \right\Vert^2_{\mathbb{H}} \right. \nonumber \\
    &\qquad \left. + \lambda_n \sum_{k \neq j} \Vert \beta_{jk} \Vert_{\text{HS}} \right\}, \nonumber
\end{align}

where \(\lambda_n\) is a tuning parameter. The non-zero elements of \(\bm{\hat{B}}_j\) indicate the functional dependencies between \(g_j\) and \(g_k\), thus defining the neighborhood \(\mathscr{N}_j\).

By repeating this procedure for each node \(j \in [p]\), we can estimate the entire edge set \(E\), capturing the spatial functional dependencies among the nodes in the graph.

\subsubsection{Temporal Functional Neighborhood Estimation}

In addition to estimating spatial functional edges, it is crucial to identify temporal dependencies within the functional graphical model. This involves detecting dependencies not only between different functions but also across time points for each function.

Let \(\bm{g}(\cdot) = (g_1(\cdot), g_2(\cdot), \ldots, g_p(\cdot))\) be a \(p\)-dimensional multivariate Gaussian process with domain \(\mathcal{T}\). For each function \(g_j\), we define its temporal neighborhood \(\mathscr{N}_j(t)\) as the set of time points \(t'\) where there is a significant temporal dependency between \(g_j(t)\) and \(g_j(t')\).

Consider the conditional expectation of \(g_j(t)\) given its values at other time points:
\[
\mathbb{E}[g_j(t) \mid g_j(t'), t' \neq t].
\]
By the Doob-Dynkin representation, there exists a measurable map \(\mathscr{B}_{jt} : \mathbb{H} \to \mathbb{H}\) such that
\[
\mathbb{E}[g_j(t) \mid g_j(t'), t' \neq t] = \mathscr{B}_{jt}(g_j(t'), t' \neq t) \quad \text{almost surely}.
\]

Assume that \(\mathscr{B}_{jt}\) belongs to the class of Hilbert-Schmidt operators \(\mathcal{B}_{\text{HS}}(\mathbb{H}, \mathbb{H})\). This assumption ensures that the infinite-dimensional nature of the functional data can be handled by finite-dimensional truncation. For each \(j \in [p]\) and \(t \in \mathcal{T}\), we have
\[
g_j(t) = \sum_{t' \neq t} \int_{\mathcal{T}} \beta_{jt}(t, t') g_j(t') \, dt' + e_{jt},
\]
where \(e_{jt} \independent g_j(t')\) for \(t' \neq t\), and \(\Vert \beta_{jt}(t, t') \Vert_{\text{HS}} < \infty\}\). Here, \(\beta_{jt}(t, t')\) represents the temporal dependency between \(g_j(t)\) and \(g_j(t')\).

The temporal neighborhood of \(g_j(t)\) is defined as:
\[
\mathscr{N}_j(t) = \{ t' \in \mathcal{T} \backslash \{t\} : \Vert \beta_{jt}(t, t') \Vert_{\text{HS}} > 0 \}.
\]

To estimate the temporal edges, we use a penalized functional regression approach. For each \(j \in [p]\) and \(t \in \mathcal{T}\), we regress \(g_j(t)\) on \(\{g_j(t') : t' \in \mathcal{T} \backslash \{t\}\}\) to estimate \(\beta_{jt}(t, t')\). The optimization problem is formulated as:
\begin{align}
    \bm{\hat{B}}_{jt} &= \arg \min_{\bm{B}_{jt}} \left\{ \frac{1}{2n} \sum_{i=1}^n \left\Vert g_{ij}(t) - \sum_{t' \neq t} \int_{\mathcal{T}} \beta_{jt}(t, t') g_{ij}(t') \, dt' \right\Vert^2_{\mathbb{H}} \right. \nonumber \\
    &\qquad \left. + \lambda_n \sum_{t' \neq t} \Vert \beta_{jt}(t, t') \Vert_{\text{HS}} \right\} \nonumber
\end{align}

where \(\lambda_n\) is a tuning parameter. The non-zero coefficients indicate significant temporal dependencies, thus defining the temporal neighborhood.

By combining the estimates of spatial and temporal neighborhoods, we obtain a comprehensive functional graphical model that captures both spatial and temporal dependencies among the functions.

\subsubsection{Vector-on-Vector Regression for Spatial and Temporal Neighborhood Selection}
\label{sec:f-on-f2v-on-v}

To estimate both spatial and temporal functional neighborhoods, we approximate the function-on-function regression problem with a finite-dimensional vector-on-vector regression problem. This method simplifies the computational complexity and makes the problem more tractable.

Given infinite-dimensional functions \(\{\bm{g_i}(\cdot)\}_{i=1}^n\), we first represent these functions using a finite \(M\)-dimensional basis. Let \(\bm{\phi_j} = \{ \phi_{jm} \}_{m=1}^\infty\) be an orthonormal basis of \(\mathbb{H}\). Using the first \(M\) basis functions, we compute projection scores for each \(k \in [p]\) and \(m \in [M]\):
\[
a_{ikm} = \langle g_{ik}, \phi_{jm} \rangle = \int_{\mathcal{T}} g_{ik}(t) \phi_{jm}(t) \, dt,
\]
and form the projection score vectors \(\bm{a_{i,k,M}} = (a_{ik1}, \dots, a_{ikM})^\top\). Thus, \(g_{ik}(\cdot) \approx \sum_{m=1}^M a_{ikm} \phi_{jm}(\cdot)\). For a fixed target node \(j\), denoted as \(p\), we represent the target function \(g_{ij}(\cdot)\) as \(g^Y_i(\cdot)\) and the other functions as \((g^{X_1}_i(\cdot), \dots, g^{X_{p-1}}_i(\cdot))^\top\). The projection scores are denoted by \(a^Y_{im}\) and \(a^{X_k}_{im}\), with vectors \(\bm{a^Y_{i,M}}\) and \(\bm{a^{X_k}_{i,M}}\). We use the notation:
\[
\bm{a^X_{i,M}} = \left( (\bm{a^{X_1}_{i,M}})^\top, \ldots, (\bm{a^{X_{p-1}}_{i,M}})^\top \right)^\top \in \mathbb{R}^{(p-1)M}.
\]

We then have the regression model:
\[
\bm{a^Y_{i,M}} = \sum_{k=1}^{p-1} \bm{B^{\ast}_{k, M}} \bm{a^{X_k}_{i,M}} + \bm{w_{i,M}} + \bm{r_{i,M}},
\]
where \(\bm{B^{\ast}_{k, M}} \in \mathbb{R}^{M \times M}\) is the regression matrix, \(\bm{w_{i,M}}\) is the noise vector, and \(\bm{r_{i,M}}\) is a bias term.

The truncated neighborhood of node \(j\) is defined as:
\[
\mathscr{N}^M_j = \left\{ k \in [p] \backslash \{j\} : \Vert \bm{B^{\ast}_{k, M}} \Vert_{\text{F}} > 0 \right\},
\]
where \(\Vert \bm{B^{\ast}_{k, M}} \Vert_{\text{F}}\) is the Frobenius norm. Given \(n\) i.i.d. samples \(\{\bm{g_i}(\cdot)\}_{i=1}^n\), we estimate \(\bm{B^{\ast}_{k, M}}\) using a penalized least squares approach:
\begin{align}
    &\bm{\hat{B}}_{1, M}, \ldots, \bm{\hat{B}}_{p-1, M} \nonumber \\
    &\in \arg \min_{\bm{B}_1, \ldots, \bm{B}_{p-1}} \left\{ \frac{1}{2n} \sum_{i=1}^n \left\Vert \bm{a}^Y_{i, M} - \sum_{k=1}^{p-1} \bm{B}_k \bm{a}^{X_k}_{i, M} \right\Vert^2_2 \right. \nonumber \\
    &\qquad \left. + \lambda_n \sum_{k=1}^{p-1} \Vert \bm{B}_k \Vert_{\text{F}} \right\} \nonumber
\end{align}

where \(\lambda_n\) is a tuning parameter.

\textbf{Temporal Neighborhoods:} To estimate temporal edges, we follow a similar procedure for each function \(g_j(t)\) across time points. We regress \(g_j(t)\) on \(g_j(t')\) for \(t' \neq t\) using the same penalized regression approach. This yields the temporal neighborhood:
\[
\mathscr{N}_j(t) = \{ t' \in \mathcal{T} \backslash \{t\} : \Vert \bm{B^{\ast}_{jt, M}} \Vert_{\text{F}} > 0 \}.
\]

\textbf{Combining Spatial and Temporal Neighborhoods:} By combining spatial and temporal neighborhood estimates, we construct a comprehensive functional graphical model. The estimated edge set \(\hat{E}\) is obtained by combining the estimated neighborhoods for each node and each time point, using either the AND or OR rule as described by Meinshausen and Bühlmann \cite{Meinshausen2006High}:

\textbf{AND}: if $j \in \hat{\mathscr{N}}_l$ and $l \in \hat{\mathscr{N}}_j \Rightarrow (j,l) \in \hat{E}$; \quad \textbf{OR}: if $j \in \hat{\mathscr{N}}_l$ or $l \in \hat{\mathscr{N}}_j \Rightarrow (j,l) \in \hat{E}$.

\subsubsection{RSNN as an Orthonormal Basis for Vector-on-Vector Regression}

A crucial aspect of our approach is the selection of the basis \(\bm{\phi_j}\). We propose using a Recurrent Spiking Neural Network (RSNN) as the basis function, specifically modeled with Leaky Integrate-and-Fire (LIF) neurons and Spike-Timing-Dependent Plasticity (STDP) synapses. This choice leverages the dynamic and temporal properties of RSNNs to capture complex patterns in the data.

For a chosen node \(j \in [p]\), and any \(i \in [n]\), suppose we have an estimate \(\{\hat{\phi}_{jm}\}_{m \geq 1}\) of the "true" basis \(\{\phi_{jm}\}_{m \geq 1}\). Let \(\hat{a}^Y_{im} = \langle g^Y_i, \hat{\phi}_{jm} \rangle\), \(\hat{a}^{X_k}_{im} = \langle g^{X_k}_{i}, \hat{\phi}_{jm} \rangle\), \(\bm{\hat{a}^Y_{i,M}} = (\hat{a}^Y_{i1}, \ldots, \hat{a}^Y_{iM})^\top\), and \(\bm{\hat{a}^{X_k}_{i,M}} = (\hat{a}^{X_k}_{i1}, \ldots, \hat{a}^{X_k}_{iM})^\top\). Therefore, we have:
\[
\bm{\hat{a}^Y_{i,M}} = \sum^{p-1}_{k=1} \bm{B^{\ast}_{k,M}} \bm{\hat{a}^{X_k}_{i,M}} + \bm{w_{i,M}} + \bm{r_{i,M}} + \bm{v_{i,M}},
\]
where the additional term \(\bm{v_{i,M}}\) arises from using \(\bm{\hat{\phi_j}}\) instead of \(\bm{\phi_j}\). When \(\bm{\hat{\phi_j}}\) is close to \(\bm{\phi_j}\), the error term \(\bm{v_{i,M}}\) should be small.

Using RSNNs as the basis functions offers several advantages. RSNNs can capture both spatial and temporal dependencies due to their recurrent nature and spiking dynamics. The LIF neurons are particularly effective for modeling the dynamics of biological neurons, while the STDP synapses allow for learning synaptic weights based on the timing of spikes, capturing temporal correlations naturally.

\textbf{Construction of RSNN Basis Functions:} When using an RSNN as the basis, we construct the basis functions \(\{\phi_{jm}\}_{m=1}^{M}\) by training the RSNN on the observed data and using the resulting spike trains to form the basis. This involves encoding the input signals into spike trains and then using the spike response of the network as the basis functions. Each function \(g_{ij}(\cdot)\) can be represented as a linear combination of these basis functions:
\[
g_{ij}(\cdot) \approx \sum_{m=1}^{M} a_{ijm} \phi_{jm}(\cdot),
\]
where \(a_{ijm}\) are the coefficients obtained by projecting \(g_{ij}(\cdot)\) onto the basis functions \(\{\phi_{jm}(\cdot)\}_{m=1}^{M}\).

The choice of an RSNN as the basis function is motivated by its ability to capture non-linear and dynamic patterns in the data that are often present in real-world signals.

\textbf{Orthogonality of Spike Trains:} To demonstrate that the RSNN can serve as an orthogonal basis function for functional principal component analysis (FPCA), we first need to establish that the spike trains generated by the network are orthogonal. Let \(s_i(t)\) denote the spike train of neuron \(i\), which can be expressed as a sum of Dirac delta functions:
\[
s_i(t) = \sum_{k} \delta(t - t_i^k),
\]
where \(t_i^k\) represents the \(k\)-th spike time of neuron \(i\). We define the inner product between two spike trains \(s_i(t)\) and \(s_j(t)\) as:
\[
\langle s_i, s_j \rangle = \int_0^T s_i(t) s_j(t) \, dt.
\]
For the spike trains to form an orthogonal basis, the inner product must satisfy
\[
\langle s_i, s_j \rangle = \delta_{ij} \cdot \| s_i \|^2,
\]
where \(\delta_{ij}\) is the Kronecker delta. This implies that for \(i \neq j\), \(\langle s_i, s_j \rangle = 0\}\). Using the properties of Dirac delta functions, we have
\[
\langle s_i, s_j \rangle = \sum_{k,l} \delta(t_i^k - t_j^l).
\]

In a recurrent SNN with STDP synapses, the temporal difference in spike times \(\Delta t\) ensures that spikes from different neurons do not coincide frequently, leading to:
\[
\sum_{k,l} \delta(t_i^k - t_j^l) \approx 0 \quad \text{for} \quad i \neq j.
\]

\textbf{Lemma:} \textit{Given the orthogonal spike trains \(\{s_i(t)\}\), we construct the covariance matrix \(\mathbf{C}\) of the spike trains:
\[
C_{ij} = \langle s_i, s_j \rangle.
\]
Since the spike trains are orthogonal, \(\mathbf{C}\) is a diagonal matrix:
\[
\mathbf{C} = \text{diag}(\| s_1 \|^2, \| s_2 \|^2, \ldots, \| s_N \|^2).
\]
The eigenvalues of \(\mathbf{C}\) represent the variance captured by each spike train, and the corresponding eigenvectors form the orthogonal basis functions for FPCA.}

\textbf{Proof:}

   Let \(s_i(t)\) and \(s_j(t)\) be the spike trains of neurons \(i\) and \(j\), respectively, expressed as sums of Dirac delta functions:
   \[
   s_i(t) = \sum_{k} \delta(t - t_i^k),
   \]
   where \(t_i^k\) represents the \(k\)-th spike time of neuron \(i\). The inner product between two spike trains \(s_i(t)\) and \(s_j(t)\) is defined as:
   \[
   \langle s_i, s_j \rangle = \int_0^T s_i(t) s_j(t) \, dt.
   \]

   The covariance matrix \(\mathbf{C}\) of the spike trains is defined by:
   \[
   C_{ij} = \langle s_i, s_j \rangle.
   \]
   Since \(s_i(t)\) and \(s_j(t)\) are sums of Dirac delta functions, we can write:
   \[
   \langle s_i, s_j \rangle = \int_0^T \left( \sum_{k} \delta(t - t_i^k) \right) \left( \sum_{l} \delta(t - t_j^l) \right) dt.
   \]
   Simplifying the integral, we get:
   \[
   \langle s_i, s_j \rangle = \sum_{k} \sum_{l} \int_0^T \delta(t - t_i^k) \delta(t - t_j^l) \, dt.
   \]
   Using the sifting property of Dirac delta functions:
   \[
   \langle s_i, s_j \rangle = \sum_{k} \sum_{l} \delta(t_i^k - t_j^l).
   \]
   
   For the spike trains to be orthogonal, the inner product \(\langle s_i, s_j \rangle\) must satisfy:
   \[
   \langle s_i, s_j \rangle = \delta_{ij} \| s_i \|^2,
   \]
   where \(\delta_{ij}\) is the Kronecker delta, which is 1 if \(i = j\) and 0 otherwise. This implies that for \(i \neq j\):
   \[
   \sum_{k} \sum_{l} \delta(t_i^k - t_j^l) = 0 \quad \text{for} \quad i \neq j.
   \]
   Therefore, the off-diagonal elements of the covariance matrix \(\mathbf{C}\) are zero:
   \[
   C_{ij} = 0 \quad \text{for} \quad i \neq j.
   \]
   For the diagonal elements where \(i = j\):
   \[
   C_{ii} = \langle s_i, s_i \rangle = \| s_i \|^2.
   \]
   Thus, the covariance matrix \(\mathbf{C}\) is a diagonal matrix:
   \[
   \mathbf{C} = \text{diag}(\| s_1 \|^2, \| s_2 \|^2, \ldots, \| s_N \|^2).
   \]

   Since \(\mathbf{C}\) is a diagonal matrix, the eigenvalues are simply the diagonal elements \(\| s_i \|^2\). The eigenvectors of a diagonal matrix are the standard basis vectors in \(\mathbb{R}^N\). Each eigenvector corresponds to one of the orthogonal spike trains. The eigenvalues \(\| s_i \|^2\) represent the variance captured by each spike train \(s_i\). In the context of FPCA, these eigenvalues indicate how much of the total variance in the data is captured by each principal component (spike train). The orthogonal spike trains form an orthonormal basis for the functional data space, and the covariance matrix \(\mathbf{C}\) being diagonal confirms this. The eigenvalues represent the captured variance, and the eigenvectors form the orthogonal basis functions for FPCA. Thus, the RSNN with STDP provides a set of orthogonal spike trains, which are used to construct the covariance matrix and perform FPCA, capturing the variance in the functional data.

In summary, using an RSNN as the basis function provides a powerful and flexible method for estimating the graph structure, capturing both spatial and temporal dependencies dynamically and efficiently.

\subsubsection{Construction of the Spatio-Temporal Functional Graph}

Combining the estimated spatial and temporal edges, we construct a comprehensive spatio-temporal functional graph $G_t = (V, E_t)$, capturing intricate neuronal dependencies. Edges are included based on both spatial and temporal correlations, providing a robust framework for understanding complex interactions in neuronal activity.

\textbf{Node and Edge Representation}
\begin{itemize}
    \item \textbf{Nodes ($V$)}: Each node in the graph represents a neuron in the RSNN.
    \item \textbf{Edges ($E_t$)}: Edges represent the functional connections between neurons, derived from both spatial and temporal correlations.
\end{itemize}

In the context of neural networks, especially spiking neural networks (SNNs), a receptive field refers to the specific region of sensory input that a particular neuron responds to. In other words, it is the subset of input features or input space that influences the activity of a given neuron.
The process of estimating edges in the spatio-temporal functional graph involves the following steps:

\textbf{RSNN for Functional Neighborhood Estimation: }We begin by training an RSNN on the input signals, ensuring that the membrane potentials of the neurons capture the underlying dynamics of the input. Let $\mathcal{T} \subseteq \mathbb{R}$ be a closed interval representing time, and let $V_i(t)$ denote the membrane potential of neuron $i$ at time $t$. After training, we compute centrality measures (e.g., eigenvector centrality) for each neuron in the RSNN. Let $C_i$ denote the centrality score of neuron $i$. We select the top $M$ neurons with the highest centrality scores as the basis neurons.

\textbf{Projection Using Membrane Potentials}: The membrane potentials of the selected central neurons form the basis functions for projecting input signals. Let $\phi_{jm}(t) = V_m(t)$ for the $m$-th central neuron. For each input signal $X_i(t)$, we compute the projection scores using the membrane potentials of the basis neurons:
\begin{equation}
    a_{ikm} = \int_{\mathcal{T}} X_i(t) \phi_{jm}(t) \, dt = \int_{\mathcal{T}} X_i(t) V_m(t) \, dt.
\end{equation}
Let $\bm{a^Y_{i, M}} = (a_{i1}, \ldots, a_{iM})^{\top}$ denote the vector of projection scores for neuron $j$ using the top $M$ basis neurons.

\textbf{Conditional Expectation Representation: }We express the membrane potential $V_j(t)$ of each neuron $j$ as a linear combination of the projection scores of the other neurons:
\begin{equation}
    V_j(t) = \sum_{k \neq j} \int_{\mathcal{T}} \beta_{jk} (t, t') V_k(t') \, dt' + e_j(t),
\end{equation}
where $\beta_{jk}(t, t')$ are the coefficients to be estimated, and $e_j(t)$ is a Gaussian error term. Assume the coefficients $\beta_{jk}(t, t')$ are Hilbert-Schmidt operators, ensuring the relationships are smooth and finite-dimensional. Expand $\beta_{jk}(t, t')$ using an orthonormal basis $\{\phi_m\}_{m=1}^{\infty}$:
\begin{equation}
    \beta_{jk} (t,t') = \sum^{\infty}_{m,m'=1} b^{\ast}_{jk,mm'} \phi_m (t) \phi_{m'} (t').
\end{equation}


We use penalized regression (group lasso) to estimate the coefficients $\beta_{jk}$:
\begin{equation}
    \bm{\hat{B}}_1, \ldots, \bm{\hat{B}}_{p-1} \in \arg \min_{\bm{B}_1, \ldots, \bm{B}_{p-1}} \left\{ \frac{1}{2n} \sum_{i=1}^n \left\|\bm{a^Y_{i, M}} - \sum^{p-1}_{k=1} \bm{B_k} \bm{a^{X_k}_{i, M}} \right\|^2_2 + \lambda_n \sum^{p-1}_{k=1} \|\bm{B_k}\|_{\text{F}} \right\},
\end{equation}
where $\bm{a^Y_{i, M}}$ and $\bm{a^{X_k}_{i, M}}$ are the projection scores. We solve the optimization problem using the Alternating Direction Method of Multipliers (ADMM):
\begin{equation}
    \min_{\bm{P}, \bm{Q}} \ \frac{1}{2n} \left\|\bm{A^Y} - \bm{A^X} \bm{Q} \right\|_{\text{F}}^2 + \lambda_n \sum^{p-1}_{k=1} \|\bm{P_k}\|_{\text{F}} \quad \text{subject to } \bm{P} - \bm{Q} = 0,
\end{equation}
where $\bm{A^Y}$ and $\bm{A^X}$ are matrices of projection scores.

\textbf{Estimating Neighborhoods:} We determine the neighborhood $\mathscr{N}_j$ for each neuron based on the estimated coefficients:
\begin{equation}
    \hat{\mathscr{N}}_j = \{ k \in [p-1] \, : \, \|\bm{\hat{B}_k}\|_F > \epsilon_n \}.
\end{equation}
Finally, we combine the estimated neighborhoods to form the overall functional graph. We use logical rules such as AND or OR to define edges in the graph:

\textbf{AND}: $\exists$ edge between $j, k$ if $j \in \hat{\mathscr{N}}_k$ and $k \in \hat{\mathscr{N}}_j$; \quad \textbf{OR}: $\exists$ edge between $j, k$ if $j \in \hat{\mathscr{N}}_k$ or $k \in \hat{\mathscr{N}}_j$.

\subsection{Dynamic Node Embedding}

SDGN calculates the node embedding matrix \( H_n \) using a spiking neural network (SNN)-based graph neural network and temporal attention on the graph sequence \(\{(G_i, X_i)\}_{i=1}^n\). The process involves two steps: (1) Spatial propagation and (2) Temporal propagation. Once \( H_n \) is obtained, encapsulating past multivariate event sequences \(\mathcal{H}_t\), it is used to predict future event times and types.

\subsubsection{Spatial Module}

The spatial module aggregates the influence of event occurrences and non-occurrences of relevant events, represented by neighboring nodes, to update the representation of the target node. For instance, at time $t_1$, the event occurrence of node 2 influences nodes 1 and 3, while non-event occurrences of nodes 1 and 3 influence node 2.

In our framework, we use a spiking neural network (SNN) based graph network to update the node embeddings with a predefined event graph. We employ the message-passing neural network (MPNN) \cite{gilmer2017neural} to build the event propagation framework. MPNN updates input node features through the message production, aggregation, and updating steps. The MPNN layer first produces messages $m_i^{uv}$ to model the hidden relational information between the source node $u$ and the destination node $v$ at $t_i$ as follows:
\begin{equation}
    m_i^{uv} = f_m(x_i^u, x_i^v; \theta_m)
\end{equation}
where $f_m$ is the spiking neural network that generates the message \cite{pfeiffer2018deep}. The generated messages are then aggregated as follows:
\begin{equation}
    \tilde{m}_i^v = \sum_{u \in \mathcal{N}(v)} \alpha_i^{uv} m_i^{uv}
\end{equation}
where $\alpha_i^{uv}$ is the attention coefficient quantifying the importance of message $m_i^{uv}$ to node $v$. $\alpha_i^{uv}$ is computed as:
\begin{equation}
    z_i^{uv} = f_\alpha(x_i^u, x_i^v; \theta_\alpha)
\end{equation}
\begin{equation}
    \alpha_i^{uv} = \frac{\exp(z_i^{uv})}{\sum_{u \in \mathcal{N}(v)} \exp(z_i^{uv})}
\end{equation}
where $f_\alpha$ is a spiking neural network \cite{lee2016training}.

The aggregated messages $\tilde{m}_i^v$ summarize the influence of events of other types on the target node $v$. $\tilde{m}_i^v$ can be considered as the localized global state information for the target node $v$ with respect to its neighboring event types. The attention mechanisms allow the differentiation of the importance of interactions between different event types.

\subsubsection{Temporal Module}
\label{sec:ta}
The temporal module propagates the impact of the previous event history into the future while accounting for the aggregated node embedding computed from the spatial module. This transition model projects past events into the future, allowing the estimation of event occurrence probability at any time in the future. Using the previous node embedding $h_{i-1}^v$, the temporal module computes the updated node embedding $h_i^v$ of the target event type $v$ as $\displaystyle h_i^v = \text{SNN}(h_{i-1}^v, r_i^v)$
where $r_i^v \in \mathbb{R}^d$ is the input feature for the temporal module. We use the Temporal Attention (TA) mechanism adapted for SNNs \cite{li2023scaling} to compute the sequence of input $R = (r_1^v, ..., r_n^v)$ as $\displaystyle
    R = \text{Softmax}\left(\frac{QK^\top}{\sqrt{d_k}}\right)V$
where $Q = \tilde{M}W_Q$, $K = \tilde{M}W_K$, and $V = \tilde{M}W_V$ are, respectively, the query, key, and value matrices. The matrix $\tilde{M}^v = [\tilde{m}_1^v, ..., \tilde{m}_n^v]^\top$ is the collection of the aggregated messages for node $v$ using the spatial module. The weight parameters $W_Q \in \mathbb{R}^{d \times n}$, $W_K \in \mathbb{R}^{d \times n}$, and $W_V \in \mathbb{R}^{d \times n}$ need to be trained for temporal representation learning. By applying the scale-dot product, the temporal module extracts the temporal dependency, paying attention to the significant events among the event stream. The Temporal Attention (TA) mechanism for SNNs ensures that the spiking temporal patterns are effectively captured and utilized for future event prediction. This adaptation leverages the precise timing capabilities of SNNs, enhancing the overall performance of the SDGN framework.

\textbf{Temporal-wise Attention for SNNs}

The purpose of the Temporal Attention (TA) module is to evaluate the saliency of each frame. This saliency score should consider not only the statistical characteristics of the input at the current timestep but also the information from neighboring frames. We implement this using a squeeze step and an excitation step in the temporal domain.

Let \( X_t^{n-1} \in \mathbb{R}^{L \times B \times C} \) represent the spatial input tensor of the \( n \)-th layer at timestep \( t \), where \( C \) denotes the channel size. The squeeze step calculates a statistical vector of event numbers. The statistical vector \( s_t^{n-1} \in \mathbb{R}^T \) at timestep \( t \) is given by:
\[
s_t^{n-1} = \frac{1}{L \times B \times C} \sum_{k=1}^{C} \sum_{i=1}^{L} \sum_{j=1}^{B} X_t^{n-1}(k, i, j)
\]

In the excitation step, \( s_t^{n-1} \) undergoes nonlinear mapping through a two-layer fully connected network to capture the correlation between different frames, resulting in the score vector:
\[
d_t^{n-1} = 
\begin{cases} 
\sigma \left( W_2^n \delta \left( W_1^n s_t^{n-1} \right) \right), & \text{training} \\
f \left( \sigma \left( W_2^n \delta \left( W_1^n s_t^{n-1} \right) \right) - d_{th} \right), & \text{inference}
\end{cases}
\]
where \( \delta \) and \( \sigma \) are ReLU and sigmoid activation functions, respectively. \( W_1^n \in \mathbb{R}^{T \times T} \) and \( W_2^n \in \mathbb{R}^{T \times T} \) are trainable parameter matrices, \( r \) is an optional parameter for controlling model complexity, and \( f(\cdot) \) is a Heaviside step function with \( d_{th} \) being the attention threshold. During training, the score vector trains a complete network. During inference, frames with scores lower than \( d_{th} \) are discarded, and the attention score of other frames is set to 1.

Using \( d_t^{n-1} \) as the input score vector, the final input at timestep \( t \) is:
\[
\widetilde{X}_t^{n-1} = d_t^{n-1} \ast X_t^{n-1}
\]
where \( \widetilde{X}_t^{n-1} \in \mathbb{R}^{L \times B \times C} \) is the input with attention scores.

The membrane potential behaviors of the TA-LIF and TA-LIAF layers follow:
\[
U_t^n = H_t^{n-1} + g \left( W^n \widetilde{X}_t^{n-1} \right)
\]

The excitation step maps the statistical vector \( z \) to a set of temporal-wise input scores. In this regard, the TA module enhances the SNN’s ability to focus on important temporal features.






\subsection{Estimating Intensity and Next Event Time}

In this step, the SDGN employs \( h_n^0 \) to estimate the intensity \( \lambda^v(t | \mathcal{H}_n^0) \) for all \( v \in \mathcal{V} \). The model presumes that \( \lambda^v(t|\mathcal{H}_n) \) depends solely on the event features from the parent nodes of \( \mathcal{G} \), specifically \( \lambda^v(t|\mathcal{H}_n) = \lambda^v(t|x_{n^v_t}^k, u \in \mathcal{N}(v)) \). The representation \( h_n^0 \) is updated using features from neighboring nodes \( \{ x_{n^u_t}^k | u \in \mathcal{N}(v) \} \) via MPNN within the spatial propagation module, effectively substituting \( \mathcal{H}_n^v \) with \( h_n^0 \).

The conditional intensity function is thus defined as:
\[
\lambda^v(t | h_n^0) = \text{Softplus}(w^v \cdot h_n^0 + \delta^v (t - t_n) + b^v)
\]
where \( w^v \in \mathbb{R}^M \) is a learnable parameter specific to \( v \), \( \delta^v \in \mathbb{R} \) acts as an event modulating parameter, and \( b^v \in \mathbb{R} \) is the base intensity function for event type \( v \).

In this formulation, the initial term encapsulates the hidden states of neighboring nodes \( \mathcal{N}(v) \) which encode past events before \( t_n \). The subsequent term captures the influence of the latest event. This influence can either magnify or attenuate the intensity of the next event, contingent on the sign of \( \delta^v \).

The likelihood function for the next event time is defined as:
\[
f(t) = \lambda^v(t|h_n^0) \exp\left( - \sum_{v} \int_{t_n}^{t} \lambda^v(\tau | h_n^0) d\tau \right)
\]

Using this likelihood function, we compute the expected value of the next event time as:
\[
\mathbb{E}[T] = \int_{t_n}^{\infty} \tau f(\tau) d\tau
\]

\subsubsection{Model Inference}

In the training phase, we optimize the model parameters for the graph recurrent neural network (GRNN) and intensity functions by maximizing the log-likelihood of the dataset \( \mathcal{D} = \{(t_i, e_i)\}_{i=1}^n \). The log-likelihood is given by:
\[
\mathcal{L}(\mathcal{D}|\theta) = \sum_i \sum_{v \in \mathcal{V}_i'} \left[ \mathbb{1}(v = e_i) \log \lambda^v(t_i | h_i^0) - \int_{t_{i-1}}^{t_i} \lambda^v(\tau | h_i^0) d\tau \right]
\]

The first term represents the probability that event \( e_i = v \) occurs at \( t_i \), and the second term accounts for the probability of no events occurring between \( [t_{i-1}, t_i] \). Here, \( \mathcal{V}_i' \) includes randomly sampled nodes and the node where \( e_i \) occurs.

To handle large event sets, we use a subset of events to avoid overfitting to non-event data. After hyperparameter tuning, we set the sample size to 8. For the integral \( \int_{t_{i-1}}^{t_i} \lambda^v(\tau | h_i^0) d\tau \), we use the Monte Carlo method, dividing the interval into 10 samples from a uniform distribution. This balances computational cost and accuracy efficiently.

\subsubsection{Optimizing Intensity Functions Using SNNs}

The optimization process involves training the SNN-based intensity functions. Given the conditional intensity function:
\[
\lambda^v(t | h_n^0) = \text{Softplus}(w^v \cdot h_n^0 + \delta^v (t - t_n) + b^v)
\]

The gradients for the parameters \( w^v \), \( \delta^v \), and \( b^v \) are computed using backpropagation through the SNN layers. The parameters are updated using gradient descent to maximize the log-likelihood function:
\[
\frac{\partial \mathcal{L}}{\partial w^v} = \sum_i \mathbb{1}(v = e_i) \frac{h_i^0}{\lambda^v(t_i | h_i^0)} - \int_{t_{i-1}}^{t_i} \frac{h_i^0}{\lambda^v(\tau | h_i^0)} d\tau
\]

\[
\frac{\partial \mathcal{L}}{\partial \delta^v} = \sum_i \mathbb{1}(v = e_i) \frac{(t_i - t_{i-1})}{\lambda^v(t_i | h_i^0)} - \int_{t_{i-1}}^{t_i} \frac{(\tau - t_{i-1})}{\lambda^v(\tau | h_i^0)} d\tau
\]

\[
\frac{\partial \mathcal{L}}{\partial b^v} = \sum_i \mathbb{1}(v = e_i) \frac{1}{\lambda^v(t_i | h_i^0)} - \int_{t_{i-1}}^{t_i} \frac{1}{\lambda^v(\tau | h_i^0)} d\tau
\]

These gradients are used to update the parameters iteratively until convergence. By training the model in this manner, the SDGN framework can effectively predict the intensity and timing of future events, leveraging the spiking neural network's ability to capture temporal dependencies and non-linear interactions.

\section{Datasets}

\label{sec:synth_data}
\begin{figure}
    \centering
    \includegraphics[width=0.8\columnwidth]{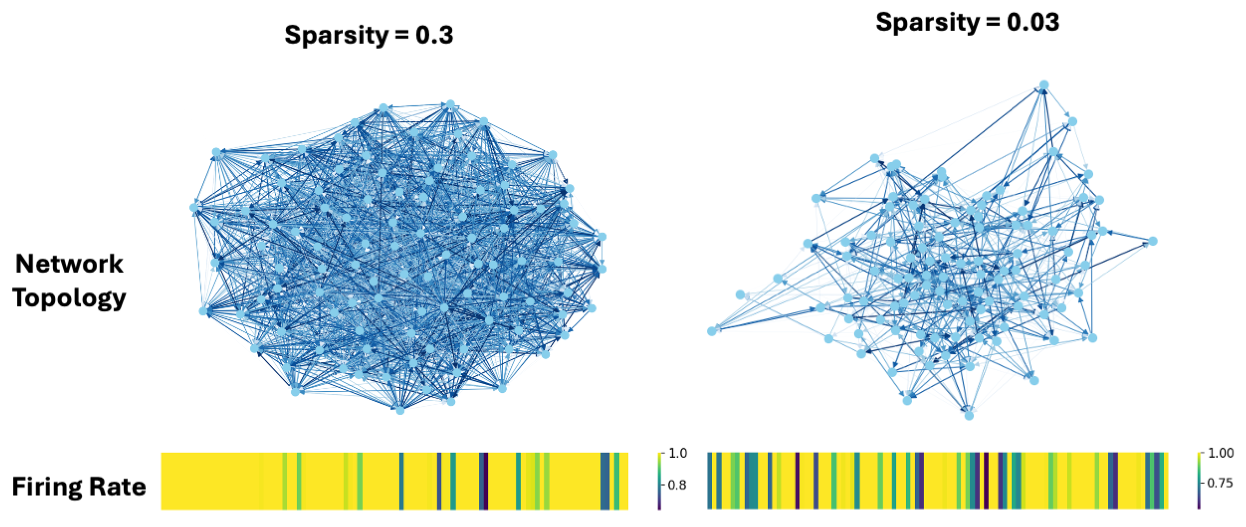}
    \caption{Visualization of the Synthetic Dataset with varying levels of sparsity and the corresponding average firing rate of the neurons}
    \label{fig:synthetic}
\end{figure}

\subsection{Synthetic Dataset for Multivariate Event Streams}

In this section, we describe the construction of a synthetic dataset designed to simulate multivariate event streams using dynamic random graphs. Each node in the graph generates events according to a Hawkes process, with the intensity function influenced by the node's connectivity. Additionally, the graph's structure evolves over time by incorporating random temporal edges.

\subsubsection{Event Stream Generation}

The event streams are generated using a Hawkes process for each node $i \in \{1, \ldots, N\}$ in the graph. The intensity function $\lambda_i(t)$ at time $t$ is defined as:

\begin{equation}
\lambda_i(t) = \mu_i + \sum_{j \in \mathcal{N}(i,t)} \alpha_{ij} \int_{0}^{t} g(t - s) dN_j(s),
\end{equation}

where:
\begin{itemize}
    \item $\mu_i$ is the base intensity of node $i$.
    \item $\mathcal{N}(i,t)$ represents the set of neighboring nodes connected to node $i$ at time $t$.
    \item $\alpha_{ij}$ is the excitation parameter representing the influence of node $j$ on node $i$.
    \item $g(t - s)$ is the decay kernel function, typically an exponential function $g(t - s) = \exp(-\beta (t - s))$ with decay rate $\beta > 0$.
    \item $N_j(s)$ is the counting process for events at node $j$ up to time $s$.
\end{itemize}

The Hawkes process captures the self-exciting nature of the events, where past events increase the likelihood of future events occurring within a short period.

\subsubsection{Dynamic Graph Structure}

The underlying graph $\mathcal{G}(t) = (\mathcal{V}, \mathcal{E}(t))$ consists of a set of nodes $\mathcal{V}$ and a time-dependent set of edges $\mathcal{E}(t)$. The graph evolves over discrete time steps $t_k, k \in \{0, 1, \ldots, T\}$, where $T$ is the total number of time steps.

At each time step $t_k$:
\begin{itemize}
    \item A subset of edges $\mathcal{E}_\text{temp}(t_{k-1}) \subset \mathcal{E}(t_{k-1})$ from the previous time step $t_{k-1}$ is randomly selected.
    \item New edges $\mathcal{E}_\text{new}(t_k)$ are added to the graph by randomly connecting pairs of nodes, ensuring the graph's connectivity and temporal dynamics.
    \item The updated edge set is $\mathcal{E}(t_k) = \mathcal{E}_\text{temp}(t_{k-1}) \cup \mathcal{E}_\text{new}(t_k)$.
\end{itemize}

This dynamic updating mechanism reflects real-world scenarios where the connectivity of nodes changes over time, introducing temporal variability.

\subsubsection{Dataset Parameters}

The synthetic dataset generation is controlled by the following parameters:

\begin{itemize}
    \item \textbf{Number of Nodes (N):} Defines the total number of nodes in the graph, $\mathcal{V}$, with $|\mathcal{V}| = N$.
    \item \textbf{Graph Sparsity (S):} Controls the density of the graph, influencing the number of edges. Sparsity is quantified by the ratio of the actual number of edges to the maximum possible number of edges in a complete graph. A higher sparsity value indicates fewer edges.
\end{itemize}

The generation process is initialized with a random graph $\mathcal{G}(0)$ with $N$ nodes and edge set $\mathcal{E}(0)$ determined by the sparsity $S$. The graph evolves over time, and the event streams are generated accordingly, providing a robust testbed for evaluating algorithms in dynamic environments.

Spike train data for each neuron is generated using a multivariate Hawkes process, a choice motivated by its ability to naturally incorporate the excitatory effects of prior spikes within and across neurons. The conditional intensity function \( \lambda_i(t) \) for neuron \( i \) is defined as:
\[ 
\lambda_i(t) = \mu_i + \sum_{j \in \text{neighbors}(i)} \alpha_{ij} \exp(-\beta_{ij}(t - t_j^{\text{last}}))
\]
Here, \( \mu_i \) denotes the baseline firing rate, drawn from a uniform distribution \( U(0.5, 1.5) \) spikes per second to introduce baseline variability. The coefficients \( \alpha_{ij} \) (interaction strength) and \( \beta_{ij} \) (decay rate) are drawn from uniform distributions \( U(0.1, 0.5) \) and \( U(1, 5) \), respectively, reflecting typical synaptic dynamics. The term \( t_j^{\text{last}} \) represents the most recent spike time of neuron \( j \) before time \( t \), ensuring the temporal dependency is captured.

For each graph configuration, we simulate a continuous spike recording for 1000 seconds. Upon collecting the spike train data, our proposed estimation algorithm is applied.  We repeat each experiment 5 times with different random seeds for graph generation. A visualization of the network topology and the event firing rate is shown in Fig. \ref{fig:synthetic}.

\subsubsection{Performance Metric: SSI}
\label{sec:ssi}
In this section, we studied how well the SDGN model can estimate the underlying graphical structure. For this, we use the synthetic dataset creation procedure as discussed above. We measure the success metric of the temporal graph estimation algorithm using the Structural Similarity Index (SSI). To compute the SSI for graphs, you define a way to measure the structural similarity between the adjacency matrices of two graphs as follows:

Let \( A \) and \( \hat{A} \) be the adjacency matrices of the true graph \( G = (V, E) \) and the estimated graph \( G' = (V, E') \), respectively.

Compute the local means, variances, and covariances of the adjacency matrices:

  \[ \textbf{Mean: } \quad 
  \mu_A = \frac{1}{|V|^2} \sum_{i=1}^{|V|} \sum_{j=1}^{|V|} A_{ij}, \quad \quad 
  \mu_{\hat{A}} = \frac{1}{|V|^2} \sum_{i=1}^{|V|} \sum_{j=1}^{|V|} \hat{A}_{ij}
  \]
  \[ \textbf{Variance: } \quad
  \sigma_A^2 = \frac{1}{|V|^2 - 1} \sum_{i=1}^{|V|} \sum_{j=1}^{|V|} (A_{ij} - \mu_A)^2; \]
  \[\quad \quad \quad \quad \quad
  \sigma_{\hat{A}}^2 = \frac{1}{|V|^2 - 1} \sum_{i=1}^{|V|} \sum_{j=1}^{|V|} (\hat{A}_{ij} - \mu_{\hat{A}})^2
  \]
  \[ \textbf{Covariance: } \quad
  \sigma_{A\hat{A}} = \frac{1}{|V|^2 - 1} \sum_{i=1}^{|V|} \sum_{j=1}^{|V|} (A_{ij} - \mu_A)(\hat{A}_{ij} - \mu_{\hat{A}})
  \]

The SSI between the adjacency matrices \( A \) and \( \hat{A} \) can be computed as:
\[
\text{SSI}(A, \hat{A}) = \frac{(2 \mu_A \mu_{\hat{A}} + C_1)(2 \sigma_{A\hat{A}} + C_2)}{(\mu_A^2 + \mu_{\hat{A}}^2 + C_1)(\sigma_A^2 + \sigma_{\hat{A}}^2 + C_2)}
\]
where \( C_1 \) and \( C_2 \) are small constants added to avoid division by zero and stabilize the division with weak denominator values. They are usually set to \( C_1 = (K_1 L)^2 \) and \( C_2 = (K_2 L)^2 \), where \( L \) is the dynamic range of the pixel values (for graphs, this could be 1 if dealing with unweighted graphs) and \( K_1 \) and \( K_2 \) are small constants (e.g., \( K_1 = 0.01 \) and \( K_2 = 0.03 \)).

The SSI value ranges from -1 to 1 where an SSI of 1 indicates perfect similarity between the structures of the two graphs.
and an SSI of -1 indicates perfect dissimilarity. This measure provides a balanced evaluation that considers the similarity in local structure, accounting for means, variances, and covariances between the graphs' adjacency matrices. The SSI for different levels of the sparsity of the underlying graph for an increasing number of nodes is shown in Figure \ref{fig:ssi}.

\subsection{Real-World Datasets}\label{sec:real_data}
We use the proposed model to analyze multivariate event sequence data across various domains. The datasets include NYC TAXI, Reddit, Earthquake, Stack Overflow, and 911 Calls as described below:
\begin{itemize}
\item \textbf{NYC TAXI:} The dataset comprises millions of taxi pickup records from 2013-2019, tagged with latitude and longitude coordinates. Each of the 299 event types corresponds to a specific NYC neighborhood, creating a graph with 299 nodes.

\item \textbf{Reddit:} From the 2 million posts in January 2014, we randomly selected 1,000 users and 100 subreddit categories, forming 100 event types and 1,000 posting sequences. Nodes represent categories of posts, and edges are defined by users’ posting histories, linking related content types such as "video" and "movie." The resulting graph has 100 nodes.

\item \textbf{Earthquake:} This dataset features data on earthquakes and aftershocks in Japan from 1990 to 2020. We assigned 162 observatories as nodes, with edges based on their geographic locations. 

\item \textbf{Stack Overflow:} Utilizing data from this question-and-answer platform, we analyzed sequences of answers and awarded badges, categorized into 22 event types across 480,413 events. 

\item \textbf{911 Calls:} The dataset includes records of emergency calls from 2019, with timestamps and caller locations. There are 69 nodes assigned to cities based on the call origins.
\end{itemize}

\section{Related Works}\label{sec:related}

Predicting future events through temporal point processes (TPPs) has been extensively studied using various approaches. Traditional parametric models, such as the Poisson process and the Hawkes process, have been foundational in modeling the occurrence of events over time. Ogata's work on the Poisson process \cite{ogata1981} and Hawkes' introduction of the self-exciting process \cite{hawkes1971} are seminal contributions that have influenced numerous subsequent studies. Despite their effectiveness in specific applications, these models often fall short in capturing the complex dependencies present in real-world data due to their reliance on predefined parametric intensity functions.

Recent advancements in deep learning have introduced neural network-based methods to address the limitations of traditional parametric models. Recurrent neural networks (RNNs) and their variants, such as long short-term memory (LSTM) networks, have been employed to model temporal dependencies in event sequences. Du et al. \cite{du2016} proposed Recurrent Marked Temporal Point Processes (RMTPP), leveraging RNNs to capture the temporal dynamics of events. RMTPP presents an RNN-based embedding to generalize various temporal point processes by extracting the patterns of time intervals and event types. Mei and Eisner \cite{mei2017} introduced Neural Hawkes Processes (NHP), integrating continuous LSTM networks with Hawkes processes to enhance the modeling of event dependencies. These methods demonstrate significant improvements in capturing temporal patterns but often struggle to incorporate the relational information among events.

The application of self-attention mechanisms and Transformer models has further advanced the field of TPPs. Zhang et al. \cite{zhang2020} utilized self-attention mechanisms to model temporal dependencies, allowing for parallel computation and capturing long-range dependencies. The Temporal Hawkes Process (THP) introduces self-attention layers to capture the importance of events for historical embeddings. However, these deep neural network (DNN)-based approaches often require substantial amounts of training data and computational resources, limiting their applicability in scenarios with limited data or real-time requirements.

Graphical event models represent another significant direction in TPP research. Marked point process problems can be effectively cast into Graphical Event Modeling (GEN) problems, where each event type is assigned to nodes connected by edges representing potential dependencies. This approach enables the discovery of causal relationships between event nodes. By defining the state of nodes based on their neighboring states, graphical models help recognize meaningful features in complex data streams, enriching the representation of latent features.

Diffusion Convolutional Recurrent Neural Networks (DCRNN) proposed by Li et al. \cite{li2017} further contribute to the modeling of spatio-temporal dependencies, particularly in traffic forecasting. By combining graph convolutional networks (GCNs) with RNNs, DCRNN captures both spatial and temporal dependencies in traffic data. Similarly, Yoon et al. \cite{yoon2023} introduced a method for learning multivariate Hawkes processes using graph recurrent neural networks (GRNNs), which effectively model complex event dependencies.

In contrast to DNN-based approaches, spiking neural networks (SNNs) offer a biologically inspired framework for processing temporal data. SNNs operate based on the precise timing of spikes, making them naturally suited for event-driven processing. Gerstner and Kistler \cite{gerstner2002} highlighted the potential of SNNs in capturing temporal dynamics through spike-timing-dependent plasticity (STDP), an unsupervised learning mechanism. Masquelier et al. \cite{masquelier2007} demonstrated the effectiveness of STDP in enabling SNNs to adapt to temporal patterns in data without the need for labeled samples.

Our work distinguishes itself by integrating SNNs with dynamic graph representations to model and predict TPPs. While previous studies have utilized SNNs for temporal data processing, our approach leverages their event-driven dynamics and STDP for the dynamic estimation of spatio-temporal functional graphs. This enables the capture of both temporal and relational dependencies among events, addressing the limitations of traditional and DNN-based methods. By constructing dynamic temporal functional graphs, our Spiking Dynamic Graph Network (SDGN) enhances the expressiveness and flexibility of TPP models, leading to superior predictive performance and a deeper understanding of complex temporal interactions in various domains.

\textbf{Spiking Neural Networks}

Spiking Neural Networks (SNNs) \cite{ponulak2011introduction} leverage bio-inspired neurons and synaptic connections, which can be trained using either unsupervised localized learning rules such as spike-timing-dependent plasticity (STDP) \cite{gerstner2002mathematical, chakraborty2023braindate} or supervised learning algorithms like surrogate gradient backpropagation \cite{neftci2019surrogate}. SNNs are increasingly recognized as a powerful, low-power alternative to deep neural networks for various machine learning tasks. By processing information using binary spike representations, SNNs eliminate the need for multiplication operations during inference. Advances in neuromorphic hardware \cite{akopyan2015truenorth, davies2018loihi, kim2022moneta} have demonstrated that SNNs can achieve significant energy savings compared to artificial neural networks (ANNs) while maintaining similar performance levels. Given their growing importance as efficient learning models, it is essential to understand and compare the representations learned by different supervised and unsupervised learning techniques. Empirical studies on standard SNNs indicate strong performance across various tasks, including spatiotemporal data classification \cite{lee2017deep, khoei2020sparnet}, sequence-to-sequence mapping \cite{chakraborty2023brainijcnn, zhang2020temporal}, object detection \cite{chakraborty2021fully, kim2020spiking}, and universal function approximation \cite{gelenbe1999function, IANNELLA2001933}. Moreover, recent research has shown that introducing heterogeneity in neuronal dynamics \cite{perez2021neural, chakraborty2023heterogeneous, chakraborty2022heterogeneous, she2021heterogeneous} can enhance the performance of SNNs to levels comparable to those of supervised learning models. Furthermore, recent studies \cite{chakraborty2024topological, chakraborty2024sparse} have explored the topological representations and pruning methodologies in SNNs, revealing that heterogeneous dynamics and task-agnostic pruning strategies significantly contribute to the efficiency and performance of these networks.

\textbf{STDP-based learning in Recurrent Spiking Neural Network}
Spike-Timing-Dependent Plasticity (STDP) \cite{gerstner1993spikes, chakraborty2021characterization} is a biologically inspired learning mechanism for recurrent Spiking Neural Networks (SNNs), relying on the precise timing of spikes for synaptic weight adjustments. STDP enables these networks to learn and generate sequences and abstract hidden states from sensory inputs, making it crucial for tasks like pattern recognition and sequence generation. For example, Guo et al. \cite{guo2021supervised} proposed a supervised learning algorithm for recurrent SNNs based on BP-STDP, optimizing structured learning. Van der Veen \cite{veen2022including} explored incorporating STDP-like behavior in eligibility propagation within multi-layer recurrent SNNs, showing improved classification performance in certain neuron models. Chakraborty et al. \cite{chakraborty2022heterogeneous} introduced a heterogeneous recurrent SNN for spatio-temporal classification, employing heterogeneous STDP with varying learning dynamics for each synapse, achieving performance comparable to backpropagation-trained supervised SNNs with reduced computation and training data requirements. Additionally, Panda et al. \cite{panda2017learning} combined Hebbian plasticity with a non-Hebbian synaptic decay mechanism in a recurrent spiking model to learn stable contextual dependencies between temporal sequences, suppress chaotic activity, and enhance sequence generation consistency. Research on topological representations \cite{chakraborty2024topological} and sparse network design \cite{chakraborty2024sparse} further emphasizes the significance of heterogeneous dynamics and novel pruning techniques in advancing the capabilities of recurrent SNNs.

\end{document}